\documentclass[11pt]{article}

\newcommand{\toreview}[1]{{\color{black}#1}}

\newcommand{\toreviewpc}[1]{{\color{black}#1}}

\usepackage[preprint]{acl}

\usepackage{times}
\usepackage{latexsym}

\usepackage[T1]{fontenc}

\usepackage[utf8]{inputenc}

\usepackage{microtype}

\usepackage{inconsolata}

\usepackage{graphicx}

\usepackage{amsmath}
\usepackage{amssymb}
\usepackage{amsthm}
\usepackage{booktabs}
\usepackage{algorithm}
\usepackage{algorithmic}
\usepackage{multirow}
\usepackage{titletoc}
\usepackage{setspace}
\usepackage{subcaption}
\usepackage{xcolor}
\newcommand{\revised}[1]{\textcolor{black}{#1}}
\usepackage[most]{tcolorbox}

\tcbset{
  samplebox/.style={
    colback=blue!2,
    colframe=blue!35!black,
    colbacktitle=blue!40!black,
    coltitle=white,
    boxrule=0.6pt,
    arc=3pt,
    left=6pt, right=6pt, top=4pt, bottom=4pt,
    fonttitle=\bfseries\small,
    fontupper=\small,
    breakable,
    enhanced
  }
}

\usepackage{listings}
\usepackage{xcolor}

\definecolor{lstBg}{RGB}{248,248,250}
\definecolor{lstFrame}{RGB}{210,210,220}
\definecolor{lstComment}{RGB}{96,112,128}
\definecolor{lstKey}{RGB}{30,90,150}
\definecolor{lstStr}{RGB}{140,60,60}
\definecolor{lstRule}{RGB}{180,80,40}

\lstdefinestyle{egbase}{
  basicstyle=\footnotesize\ttfamily,
  backgroundcolor=\color{lstBg},
  frame=single,
  rulecolor=\color{lstFrame},
  framesep=4pt,
  framexleftmargin=0pt,
  framexrightmargin=0pt,
  xleftmargin=4pt,
  xrightmargin=4pt,
  linewidth=\linewidth,
  resetmargins=true,
  breaklines=true,
  breakatwhitespace=true,
  postbreak=\mbox{\textcolor{lstFrame}{$\hookrightarrow$}\space},
  columns=fullflexible,
  keepspaces=true,
  showstringspaces=false,
  literate={—}{{\textemdash}}1, 
  commentstyle=\color{lstComment}\itshape,
  captionpos=b,
  abovecaptionskip=2pt,
  belowcaptionskip=2pt,
  aboveskip=4pt,
  belowskip=4pt,
}

\lstdefinestyle{prompt}{
  style=egbase,
  language=,
  morecomment=[l]{\#},
  emph={WORKFLOW,KEY,RULES,IMPORTANT,CRITICAL,FIRST,NEVER,BASIC,ECONOMY},
  emphstyle=\color{lstRule}\bfseries,
}

\lstdefinestyle{json}{
  style=egbase,
  language=,
  morecomment=[l]{\#},
  string=[b]",
  stringstyle=\color{lstStr},
  morekeywords={input,subgoal,metadata,subgoals,gradingNotes},
  keywordstyle=\color{lstKey}\bfseries,
}

\title{\textsc{EdgeGen}: Improving Tool-Calling Agents Beyond Happy Paths with Synthetic Edge Case Generation}

\author{
  \textbf{Harshavardhan Abichandani, Penny Chong, Jiyuan Shen,} \\
  \textbf{Gunraj Singh, Ashutosh Hathidara, Marcus Duigan Xing Yu,} \\
  \textbf{Jane Lo, Atin Ghosh, Yipeng Li, Daniel Dahlmeier} \\[0.5em]
  SAP \\[0.3em]
  {\small\texttt{\{harshavardhan.abichandani,penny.chong,jiyuan.shen\}@sap.com}} \\
  {\small\texttt{\{gunraj.singh,ashutosh.hathidara,marcus.duigan,jane.lo\}@sap.com}} \\
  {\small\texttt{\{atin.ghosh,yipeng.li,d.dahlmeier\}@sap.com}}
}

\begin{document}
\maketitle
\begin{abstract}
\toreview{Tool-calling LLM agents are increasingly deployed in enterprise applications. However, effective evaluation and optimization require high-quality, diverse task datasets that are often difficult to obtain due to privacy and other constraints. Existing synthetic task generation methods often produce generic tasks that ignore an agent's underlying state or database and fail to reflect real-world usage diversity.
We propose \textsc{EdgeGen}, a synthetic task generation framework that extracts compliance rules from an agent’s specification and uses them to generate database-grounded edge-case tasks designed to violate these rules. When combined with existing synthetic data generation techniques, \textsc{EdgeGen} enables agent improvement through finetuning and harness optimization. The resulting pipeline forms a fully automated closed-loop system that requires no human annotation. Finetuning on data generated by \textsc{EdgeGen} yields a consistent mean progress improvement of \textbf{+2\%} to \textbf{+42\%} on $\tau^2$-bench airline domain, while other baseline methods show degradation for some models. On the other hand, for harness optimization, our method shows a mean progress improvement of  \textbf{+10\%} and \textbf{+30\%}  over the human-curated and base harnesses, respectively, for the Gemma-4-e4b model.

}
\end{abstract}

\section{Introduction}
\label{sec:introduction}
\begin{figure*}[t]
\centering
\includegraphics[width=0.9\textwidth]{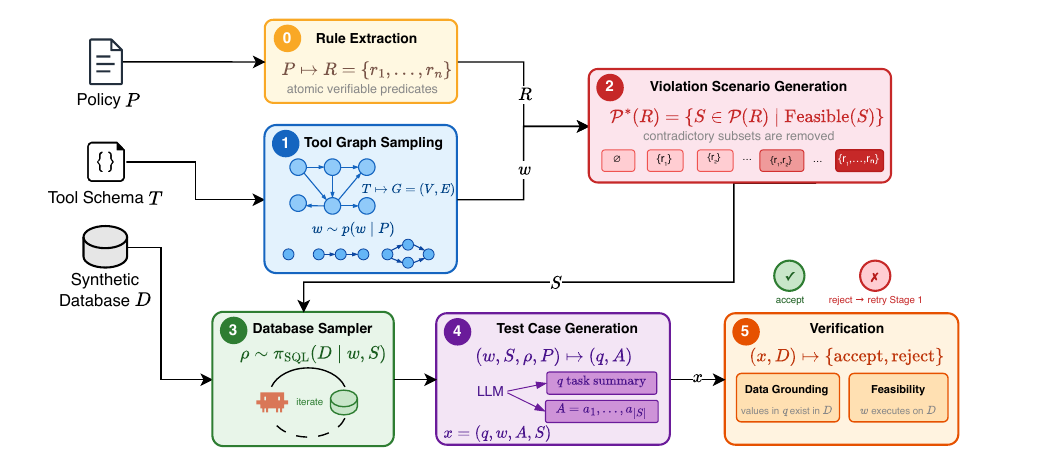}
\caption{Illustrates the automated \textsc{EdgeGen} pipeline for generating synthetic, database-grounded test cases $x$ designed to violate business policies. The system first extracts compliance rules $R$ from an agent's policy document and enumerates combinations of these rules $\mathcal{P^*}(R)$ to create structured violation scenarios $S$. It then samples a tool workflow $w$, grounds the scenario in a database $D$ state using a SQL agent $\pi_{\text{SQL}}$, and verifies the final task for data grounding and scenario feasibility.}
\label{fig:concept-diagram}
\end{figure*}

\toreview{
Large Language Model (LLM) agents are rapidly becoming a core component of enterprise systems. These agents must autonomously interact with tools, query databases, and execute multi-step workflows while adhering to complex business policies and operational constraints \citep{schick2023toolformer, patil2023gorilla, qin2024tool}. As these agents scale up and interact in more varied conditions, improving the reliability and policy compliance of such systems has become increasingly important. Whether through reinforcement learning \citep{feng2024agile, chen2025reinforcementlearninglonghorizoninteractive}, prompt engineering or harness optimization \citep{lee2026meta, zhang2026darwingodelmachineopenended, zhang2026agenticcontextengineeringevolving}, processes ultimately depend on high quality task data.

However, obtaining such data remains a major challenge. Real production logs are often inaccessible due to privacy concerns, while manually creating realistic tasks the agent might encounter is expensive and difficult to scale. At the same time, prior works on synthetic data generation consistently identify \emph{quality} and \emph{diversity} as the key factors of downstream performance \citep{liu2024best, chen2024diversity}. These limitations have motivated a growing body of work on synthetic task generation for tool-use LLM agents \citep{wang2023selfinstruct, liu2025toolace, liu2024apigen, prabhakar2026apigen}.

Despite the progress, existing approaches largely focus on generic function calling tasks and often fail to capture the operational constraints of the agents. In practice, correctness depends not only on whether the agent can invoke the right tool, but also whether it respects domain-specific business rules. For example, an airline-support agent should deny refund requests that violate eligibility constraints, apply baggage policies conditioned on loyalty tiers, etc. These are inherently stateful and policy-dependent behaviors that generic synthetic tasks rarely evaluate.

Coverage is another dimension that is equally important. Agents that perform well on narrow or templated evaluations may still fail in deployment when faced with rare edge cases \citep{zhang2024agent, ruan2024identifyingriskslmagents, zhou2024webarenarealisticwebenvironment}. Robust evaluation therefore requires diverse, grounded scenarios that systematically probe an agent's failure modes rather than testing common ``happy-path'' interactions. \revised{The distinction is not only how to execute a workflow, but when it should be refused. For example, the same cancellation request may require approval or denial depending on the database state. Informative boundary cases therefore pair a request with a consistent state that makes the expected response verifiable.}

To address this gap, we introduce \textsc{EdgeGen}, a framework for generating diverse, database-grounded tasks that explicitly stress-test business-rule compliance. Our pipeline in Figure \ref{fig:concept-diagram} first extracts these rules from an agent's policy document. It then constructs violation scenarios by enumerating combinations of these rules, grounds each scenario in database states using a scenario-aware SQL agent which samples values that make the task feasible, and verifies the validity of every generated task before admission. The resulting dataset spans a broad spectrum of difficulty, ranging from simple single tool interaction to complex multi-step workflows involving multiple policy violations.

In our setting, we pair \textsc{EdgeGen} with a synthetic database generation method \citep{anonymous2026synthesis}\footnote{Anonymous submission will be made public at a later date.} that produces a relational database that inherently encodes the business rules. Together, these components form a fully synthetic closed loop \toreview{agent} improvement pipeline. The synthetic database provides the environment state, \textsc{EdgeGen} generates grounded tasks against that state, and the resulting trajectories are used for finetuning or harness optimization. The entire process operates without production logs, human annotated tasks or real data, yet still improves performance on \toreview{human-curated} test sets.

Empirically, using supervised-finetuning and harness optimization, we show that, \textsc{EdgeGen} consistently improves performance across multiple benchmarks, model families and against different synthetic generation baselines. On $\tau^2$-bench airline \citep{barres2025tau2benchevaluatingconversationalagents} and ToolSandbox \citep{lu2024toolsandbox} it yields a substantial gain for small and mid-sized models, showing a $+42\%$ improvement on average progress for Qwen 2.5-3b relative to the base version. \toreview{For harness optimization, our method shows a relative mean progress improvement of  $+10\%$ and $+30\%$  over the human-curated and base harnesses, respectively, on $\tau^2$-bench airline domain using the Gemma-4-e4b model.}

\paragraph{Contributions}
\begin{enumerate}
    \item We propose \textsc{EdgeGen}, a violation-driven synthetic task generation framework that extracts compliance rules from agent specifications, enumerates structured combinations of rule violations, and grounds tasks in executable database states.

    \item We introduce a fully automated closed-loop system that combines synthetic database generation, \textsc{EdgeGen} task construction, produce high-quality training and evaluation data without human annotations or production logs, enabling agent optimization.

    \item We demonstrate empirically that explicitly covering edge cases improves tool-using agent performance across benchmarks and model scales.
    
\end{enumerate}
}
\section{Related Work}
\label{sec:related_work}

\subsection{Synthetic Data and Task Generation}

\toreviewpc{Self-instruct~\citep{wang2023selfinstruct} establishes the recipe of bootstrapping
\toreviewpc{instruction data from 
their own generations. Subsequent work \citet{chen2024diversity} has studied impact of synthetic data, including the role of diversity in downstream performance during pretraining and finetuning}. 
\toreviewpc{APIGen~\citep{liu2024apigen} proposes an automated pipeline for generating diverse and verifiable function-calling datasets from executable APIs, while  APIGen-MT~\citep{prabhakar2026apigen} further extends this framework to multi-turn tool-use. ToolACE~\citep{liu2025toolace} further improves synthetic function-calling data generation via self-evolving synthesis and a complexity-controlled dialog generation with verification layer to ensure data quality. 
} 

In this work, we repurpose TaskBench \citep{shen2023taskbench}, from which we adopt the tool graph sampling strategies (Node, Chain and Directed-Acyclic Graph (DAG)) for workflow construction. Similarly, we adapt FuncBenchGen \citep{maekawa2026towards} as described in Sec. ~\ref{sec:experiments}. Both share a common limitation when applied to agent benchmarks that operate using a specific behavioral policy. For TaskBench \citep{shen2023taskbench}, task descriptions are either generated using an LLM using back-instruct and sampled subgraphs. FuncBenchGen's \citep{maekawa2026towards} tasks are constructed from function schemas with programmatically assigned values. As a result, they are not systematically enumerating over a structured space of behavioral rule violations nor grounded in a database state. This leaves edge case scenarios underexplored. \textsc{EdgeGen} addresses these shortcomings by first enumerating over the power-set of behavioral rules, and second, uses a scenario-aware SQL agent to fetch actual database state.


}

\subsection{Agent Evaluation}

\toreview{
The $\tau$-Bench~\citep{yao2024taubench} and $\tau^{2}$-Bench~\citep{barres2025tau2benchevaluatingconversationalagents}, introduce realistic multi-turn customer-service scenarios, grounded in an airline or retail policy, acting as our primary evaluation benchmark.
ToolSandbox~\citep{lu2024toolsandbox} provides a tool device-management
environment with stateful constraints and minimal system prompt,
serving as our secondary benchmark. \toreviewpc{Other broader family of agent benchmark and evaluation approaches are AgentBench~\citep{liu2023agentbench},
WorkArena~\citep{drouin2024workarena}, API-Bank~\citep{li2023apibank},
MINT~\citep{wang2024mint}, and AgentBoard~\citep{ma2024agentboard} which follow common evaluation paradigm for tool-use agents.} 
\toreviewpc{While each provides a curated set of test cases, none provides a synthetic-data layer for automatically expanding existing test cases evaluation to support agent improvement.}
\textsc{EdgeGen} closes this gap by introducing a fully automated closed-loop system: from simulating compliance-aware databases and generating grounded training and evaluation tasks, to optimizing agents based on the synthetic data.}

\section{Methodology}
\label{sec:method}

\subsection{Problem Formulation}
\label{sec:problem-formulation}

\toreview{
A deployment of a tool-calling agent can be characterized by three artefacts that are typically available during development. The first is the set of tools accessible to the agent, denoted as:
$$T=\{t_1, \ldots, t_n\},$$
where each tool $t_i$ is associated with input $\text{In}(t_i)$ and output $\text{Out}(t_i)$ signatures that define its data types. The second artefact is the policy document $P$, which specifies the behavioral constraints and operational requirements the agent must follow. The third is the relational database $D$ that the agent interacts with during execution through reads and writes.

The policy document $P$ encodes behavioral rules in natural language. We formalize these rules as a set of atomic compliance constraints implied by $P$:
$$R=\{r_1, \ldots, r_n\}.$$
Each rule $r_i$ corresponds to a single verifiable assertion about the agent's behavior. For example, a rule may specify that ``the agent must verify the user identity before modifying a reservation.'' Our framework extracts these rules automatically from $P$. \revised{For scenario enumeration, we limit $R$ to the five most important extracted rules, giving at most $2^5=32$ subsets before feasibility filtering. The extraction prompt and rule-extraction analysis are provided in Appendices~\ref{app:rule-extraction-prompt} and~\ref{app:rule-extraction-audit}.}

We define a test case as a four-element tuple:
$$x=\langle q, w, A, S \rangle,$$
where $q$ denotes a user task description grounded in concrete values sampled from the database. The workflow:
$$w = \{t_i, \ldots, t_k\},$$
represents an ordered sequence of $k$ tool invocations drawn from $T$. The third element, 
$$A = \{a_1, \ldots, a_n\},$$
is the set of natural language assertions \citep{chong2026talk} associated with the task. These assertions serve as evaluation targets and are used to determine whether the agent successfully completed the task. Finally, $S \subseteq R$ denotes the set of policy rules that the test case intends to violate or stress-test.

\revised{For example, a retail task asks to exchange a delivered water bottle for a hoodie. Its assertions require authentication and denial of the cross-product-type exchange (Appendix~\ref{appendix_samples_synthesized}). The user request challenges the policy; the expected agent response remains compliant.}

A test case $x$ is considered valid with respect to the database $D$ when two conditions are satisfied. \revised{First, identifiers and database facts used to ground $q$ and $w$ must be supported by $D$; deliberately false user claims and values to be created need not already hold in $D$ (data grounding). Second, the available tools and database state must permit the expected policy-compliant response, including denial of a prohibited operation (scenario feasibility).} These validity requirements address two common failure modes observed in prior synthetic task generation methods \citep{shen2023taskbench, maekawa2026towards}: hallucinated identifiers and infeasible tool sequences whose required database state does not exist. Refer to Appendix-\ref{appendix:hallucinated-samples} for some examples.}


\paragraph{\revised{Database generation.}}
\revised{The synthetic database is populated by a Generalist Populator, an LLM agent that invokes API write operations rather than directly editing database rows \citep{anonymous2026synthesis}. Each operation's preconditions are checked against the current database state; for example, creating a reservation may require an existing user and flight. Rejected operations leave the state unchanged. Repeated runs produce a coherent relational database. These checks concern API-enforced preconditions, rather than all behavioral policies. \textsc{EdgeGen} is agnostic to the database source and can use any compatible relational database.}

\subsection{Test Case Generation Pipeline}

Based on this problem formulation, \textsc{EdgeGen} generates valid, database-grounded test cases through a multi-stage pipeline that progressively constructs workflows, enumerates policy violations, and grounds the resulting scenario in feasible database states.

\paragraph{Stage 1: Tool graph construction and path sampling.} Inspired by \citet{shen2023taskbench}, we construct a directed dependency graph $G=(V, E)$ where $V=T$ and an edge exists between two tools if their data types are compatible:
$$(t_i, t_j) \in E \iff \text{Out}(t_i) \cap \text{In}(t_j) \neq \emptyset .$$
This construction follows the resource type formulation mentioned in \citep{shen2023taskbench}. However, we identified that sampling a feasible workflow should also depend on policy $P$. Most of these agent prompts or policies mention constraints about which tool to call first before proceeding with an action. These rules are often too complex to explicitly represent them in code. Therefore, we define a distribution over valid workflow structure induced by $P$. We use an LLM to select a valid set conditioned on $P$. A workflow $w$ is sampled as:
$$w \sim p(w \mid P).$$
From this we sample a \textsc{Node}, \textsc{Chain}, or \textsc{DAG}, which correspond to a single-tool execution, linear compositions, or branching multi-tool workflows, respectively.

\paragraph{Stage 2: Violation scenario generation.}
Given the extracted rule set $R$, we define the space of all possible compliance conditions as its power set:
$$S \subseteq R, \quad S \in \mathcal{P}(R).$$
where $\mathcal{P}(R)$ specifies the power set of all the extracted rules. Each subset $S$ represents a set of simultaneously enforced or violation constraints. The size of the violation set, $|S|$, controls the complexity of the generated scenario. When $|S| = 0$, the task corresponds to a standard ``happy-path'' interaction. When $|S| = 1$, the test case isolates a single policy constraint. Larger violation sets ($|S| \geq 2$) generate more challenging scenarios that require the agent to simultaneously enforce multiple interacting rules.

However, not all subsets are \toreview{jointly satisfiable}. Therefore, we define a feasibility filter:
$$\mathcal{P}^*(R) = \{S \in \mathcal{P}(R) \mid \mathrm{Feasible}(S)\},$$
where $\mathrm{Feasible}(S)$ is implemented using an LLM-based consistency checker that removes logically incompatible rule combinations.

\paragraph{Stage 3: Scenario-aware database sampling.}

Given a workflow $w \sim p(w \mid P)$ and a violation set $S \in \mathcal{P}^*(R)$, we must instantiate a concrete database state that satisfies both the execution structure and the constraint configuration.

We define a grounding procedure implemented by an iterative SQL agent:
$$\rho \sim \pi_{\mathrm{SQL}}(D \mid w, S),$$
where $\rho$ is a set of database rows that instantiate the scenario. If no valid $\rho$ is found, the system resamples $w$ and $S$.

\paragraph{Stage 4: Test case generation.}

Given an agent's specifications $P$, sampled workflow $w \sim p(w \mid P)$, a violation set $S \in \mathcal{P}^*(R)$, and a grounded database state $\rho \subset D$, we generate the task summary $q$ and natural language assertions $A$ using an LLM. The task summary will contain concrete entities present in the database (eg: identifiers, timestamps, etc), and the natural language assertions will encode the expected agent response under the given constraints (eg: denial of an operation due to policy violation). 

\paragraph{Stage 5: Verification.}

Since both workflow sampling and LLM-based generation may introduce inconsistencies, we apply a verification step that \revised{checks} logical consistency with the database state $D$. We define a verifier, which will perform two independent checks, the first is data grounding, \revised{where the identifiers and database facts supporting the scenario are checked against $D$, including any mismatch between a user claim and stored state.} This is validated by the SQL agent, which will execute deterministic SQL queries over the database schema. The second is the feasibility of the scenario, \revised{where we check that the expected response is achievable with the available tools and state. For operations that create records, the prerequisites must exist; for denial scenarios, the state must support the reason for denial rather than permit the prohibited operation.} A test case $x$ is accepted only if both conditions hold. Otherwise, it is discarded, and the pipeline restarts from Stage 1 with a newly sampled workflow and violation set. \revised{These checks do not guarantee that every extracted rule or assertion is semantically correct (Appendix~\ref{app:rule-extraction-audit}).}

\subsection{Complexity-Aware Generation}

The generation process has two orthogonal axes of controllable complexity. The first \toreview{axis} controls the structural complexity via the workflow distribution. Increasing the sampling rate for \textsc{Chain} or \textsc{DAG} increases the complexity of the task. The second axis is scenario complexity as determined by $|S|$. Together, these \toreview{two} axes define a structured coverage space over task difficulty. Sampling uniformly or non-uniformly over this grid induces a controllable distribution over scenario complexity, spanning simple database lookups to multi-step, multi-policy decision workflows. We empirically analyze the effect of finetuning with different scenario complexities in Section~\ref{subsec:complexity-ablation}.

\section{Experiments}
\label{sec:experiments}

  \begin{table*}[t]
  \centering
  \resizebox{\textwidth}{!}{%
  \begin{tabular}{l|cccccc}
  \toprule
  \textbf{Model} & \textbf{Base} & \textbf{Human-Curated} & \textbf{Naive} & \textbf{FuncBenchGen} & \textbf{TaskBench} & \textbf{EdgeGen (Ours)} \\
  \midrule
  \multicolumn{7}{c}{\textit{$\tau^2$-bench Airline Domain}} \\
  \midrule
  Qwen2.5-3b & 0.35 \textbar{} 0.26 & \textbf{0.39}{\scriptsize\textcolor{blue}{$\uparrow$0.04}} \textbar{} 0.30{\scriptsize\textcolor{blue}{$\uparrow$0.04}} &
  0.34{\scriptsize\textcolor{red}{$\downarrow$0.01}} \textbar{} 0.29{\scriptsize\textcolor{blue}{$\uparrow$0.03}} & \textbf{0.39}{\scriptsize\textcolor{blue}{$\uparrow$0.04}} \textbar{}
  0.34{\scriptsize\textcolor{blue}{$\uparrow$0.08}} & 0.28{\scriptsize\textcolor{red}{$\downarrow$0.07}} \textbar{} 0.25{\scriptsize\textcolor{red}{$\downarrow$0.01}} &
  \textbf{0.39}{\scriptsize\textcolor{blue}{$\uparrow$0.04}} \textbar{} \textbf{0.37}{\scriptsize\textcolor{blue}{$\uparrow$0.11}} \\
  Qwen3.5-4b & 0.89 \textbar{} 0.76 & 0.90{\scriptsize\textcolor{blue}{$\uparrow$0.01}} \textbar{} 0.79{\scriptsize\textcolor{blue}{$\uparrow$0.03}} &
  \textbf{0.92}{\scriptsize\textcolor{blue}{$\uparrow$0.03}} \textbar{} 0.79{\scriptsize\textcolor{blue}{$\uparrow$0.03}} & 0.85{\scriptsize\textcolor{red}{$\downarrow$0.04}} \textbar{}
  0.72{\scriptsize\textcolor{red}{$\downarrow$0.04}} & 0.87{\scriptsize\textcolor{red}{$\downarrow$0.02}} \textbar{} 0.74{\scriptsize\textcolor{red}{$\downarrow$0.02}} &
  \textbf{0.92}{\scriptsize\textcolor{blue}{$\uparrow$0.03}} \textbar{} \textbf{0.79}{\scriptsize\textcolor{blue}{$\uparrow$0.03}} \\
  Qwen3.5-9b & 0.92 \textbar{} 0.80 & 0.87{\scriptsize\textcolor{red}{$\downarrow$0.05}} \textbar{} 0.78{\scriptsize\textcolor{red}{$\downarrow$0.02}} &
  0.92{\scriptsize\textcolor{gray}{$\pm$0.00}} \textbar{} 0.81{\scriptsize\textcolor{blue}{$\uparrow$0.01}} & 0.92{\scriptsize\textcolor{gray}{$\pm$0.00}} \textbar{}
  0.78{\scriptsize\textcolor{red}{$\downarrow$0.02}} & 0.89{\scriptsize\textcolor{red}{$\downarrow$0.03}} \textbar{} 0.78{\scriptsize\textcolor{red}{$\downarrow$0.01}} &
  \textbf{0.95}{\scriptsize\textcolor{blue}{$\uparrow$0.03}} \textbar{} \textbf{0.85}{\scriptsize\textcolor{blue}{$\uparrow$0.05}} \\
  Qwen3.5-35b & \textbf{0.92} \textbar{} 0.81 & 0.88{\scriptsize\textcolor{red}{$\downarrow$0.04}} \textbar{} 0.76{\scriptsize\textcolor{red}{$\downarrow$0.05}} &
  0.84{\scriptsize\textcolor{red}{$\downarrow$0.08}} \textbar{} 0.73{\scriptsize\textcolor{red}{$\downarrow$0.08}} & \textbf{0.92}{\scriptsize\textcolor{gray}{$\pm$0.00}} \textbar{}
  0.79{\scriptsize\textcolor{red}{$\downarrow$0.02}} & 0.85{\scriptsize\textcolor{red}{$\downarrow$0.07}} \textbar{} 0.76{\scriptsize\textcolor{red}{$\downarrow$0.05}} &
  \textbf{0.92}{\scriptsize\textcolor{gray}{$\pm$0.00}} \textbar{} \textbf{0.82}{\scriptsize\textcolor{blue}{$\uparrow$0.01}} \\
  Gemma-4-e4b & 0.66 \textbar{} 0.43 & 0.69{\scriptsize\textcolor{blue}{$\uparrow$0.03}} \textbar{} 0.50{\scriptsize\textcolor{blue}{$\uparrow$0.07}} &
  0.70{\scriptsize\textcolor{blue}{$\uparrow$0.04}} \textbar{} 0.48{\scriptsize\textcolor{blue}{$\uparrow$0.05}} & 0.69{\scriptsize\textcolor{blue}{$\uparrow$0.03}} \textbar{}
  0.51{\scriptsize\textcolor{blue}{$\uparrow$0.08}} & 0.70{\scriptsize\textcolor{blue}{$\uparrow$0.04}} \textbar{} 0.55{\scriptsize\textcolor{blue}{$\uparrow$0.12}} &
  \textbf{0.79}{\scriptsize\textcolor{blue}{$\uparrow$0.13}} \textbar{} \textbf{0.56}{\scriptsize\textcolor{blue}{$\uparrow$0.13}} \\
  Gemma-4-e2b & 0.47 \textbar{} 0.38 & 0.39{\scriptsize\textcolor{red}{$\downarrow$0.08}} \textbar{} 0.33{\scriptsize\textcolor{red}{$\downarrow$0.05}} &
  0.47{\scriptsize\textcolor{gray}{$\pm$0.00}} \textbar{} 0.40{\scriptsize\textcolor{blue}{$\uparrow$0.02}} & 0.49{\scriptsize\textcolor{blue}{$\uparrow$0.02}} \textbar{}
  0.41{\scriptsize\textcolor{blue}{$\uparrow$0.03}} & \textbf{0.56}{\scriptsize\textcolor{blue}{$\uparrow$0.09}} \textbar{} 0.39{\scriptsize\textcolor{blue}{$\uparrow$0.01}} &
  0.55{\scriptsize\textcolor{blue}{$\uparrow$0.08}} \textbar{} \textbf{0.42}{\scriptsize\textcolor{blue}{$\uparrow$0.04}} \\
  \midrule
  \multicolumn{7}{c}{\textit{ToolSandbox}} \\
  \midrule
  Qwen2.5-3b & 0.84 \textbar{} 0.60 & 0.81{\scriptsize\textcolor{red}{$\downarrow$0.03}} \textbar{} 0.64{\scriptsize\textcolor{blue}{$\uparrow$0.04}} &
  0.83{\scriptsize\textcolor{red}{$\downarrow$0.01}} \textbar{} 0.61{\scriptsize\textcolor{blue}{$\uparrow$0.01}} & \textcolor{gray}{N/A} & 0.81{\scriptsize\textcolor{red}{$\downarrow$0.03}}
  \textbar{} 0.60{\scriptsize\textcolor{gray}{$\pm$0.00}} & \textbf{0.86}{\scriptsize\textcolor{blue}{$\uparrow$0.02}} \textbar{} \textbf{0.67}{\scriptsize\textcolor{blue}{$\uparrow$0.07}}
  \\
  Qwen3.5-4b & 0.66 \textbar{} 0.55 & 0.75{\scriptsize\textcolor{blue}{$\uparrow$0.09}} \textbar{} \textbf{0.68}{\scriptsize\textcolor{blue}{$\uparrow$0.13}} &
  0.78{\scriptsize\textcolor{blue}{$\uparrow$0.12}} \textbar{} 0.67{\scriptsize\textcolor{blue}{$\uparrow$0.12}} & \textcolor{gray}{N/A} &
  \textbf{0.78}{\scriptsize\textcolor{blue}{$\uparrow$0.12}} \textbar{} 0.63{\scriptsize\textcolor{blue}{$\uparrow$0.08}} & 0.77{\scriptsize\textcolor{blue}{$\uparrow$0.11}} \textbar{}
  0.66{\scriptsize\textcolor{blue}{$\uparrow$0.11}} \\
  Qwen3.5-9b & 0.73 \textbar{} 0.60 & 0.78{\scriptsize\textcolor{blue}{$\uparrow$0.05}} \textbar{} 0.68{\scriptsize\textcolor{blue}{$\uparrow$0.08}} &
  0.78{\scriptsize\textcolor{blue}{$\uparrow$0.05}} \textbar{} 0.68{\scriptsize\textcolor{blue}{$\uparrow$0.08}} & \textcolor{gray}{N/A} &
  \textbf{0.83}{\scriptsize\textcolor{blue}{$\uparrow$0.10}} \textbar{} 0.69{\scriptsize\textcolor{blue}{$\uparrow$0.09}} & 0.82{\scriptsize\textcolor{blue}{$\uparrow$0.09}} \textbar{}
  \textbf{0.71}{\scriptsize\textcolor{blue}{$\uparrow$0.11}} \\
  Qwen3.5-35b & 0.70 \textbar{} 0.54 & \textbf{0.81}{\scriptsize\textcolor{blue}{$\uparrow$0.11}} \textbar{} 0.65{\scriptsize\textcolor{blue}{$\uparrow$0.11}} &
  0.76{\scriptsize\textcolor{blue}{$\uparrow$0.06}} \textbar{} 0.59{\scriptsize\textcolor{blue}{$\uparrow$0.05}} & \textcolor{gray}{N/A} & 0.73{\scriptsize\textcolor{blue}{$\uparrow$0.03}}
  \textbar{} 0.63{\scriptsize\textcolor{blue}{$\uparrow$0.09}} & \textbf{0.81}{\scriptsize\textcolor{blue}{$\uparrow$0.11}} \textbar{} \textbf{0.65}{\scriptsize\textcolor{blue}{$\uparrow$0.11}} \\
  Gemma-4-e4b & 0.82 \textbar{} 0.64 & 0.88{\scriptsize\textcolor{blue}{$\uparrow$0.06}} \textbar{} 0.76{\scriptsize\textcolor{blue}{$\uparrow$0.12}} &
  \textbf{0.92}{\scriptsize\textcolor{blue}{$\uparrow$0.10}} \textbar{} 0.77{\scriptsize\textcolor{blue}{$\uparrow$0.13}} & \textcolor{gray}{N/A} &
  0.88{\scriptsize\textcolor{blue}{$\uparrow$0.06}} \textbar{} 0.74{\scriptsize\textcolor{blue}{$\uparrow$0.10}} & 0.90{\scriptsize\textcolor{blue}{$\uparrow$0.08}} \textbar{}
  \textbf{0.77}{\scriptsize\textcolor{blue}{$\uparrow$0.13}} \\
  Gemma-4-e2b & 0.65 \textbar{} 0.53 & 0.78{\scriptsize\textcolor{blue}{$\uparrow$0.13}} \textbar{} \textbf{0.65}{\scriptsize\textcolor{blue}{$\uparrow$0.12}} &
  0.73{\scriptsize\textcolor{blue}{$\uparrow$0.08}} \textbar{} 0.62{\scriptsize\textcolor{blue}{$\uparrow$0.09}} & \textcolor{gray}{N/A} & 0.78{\scriptsize\textcolor{blue}{$\uparrow$0.13}}
  \textbar{} 0.63{\scriptsize\textcolor{blue}{$\uparrow$0.10}} & \textbf{0.85}{\scriptsize\textcolor{blue}{$\uparrow$0.20}} \textbar{} \textbf{0.65}{\scriptsize\textcolor{blue}{$\uparrow$0.12}} \\
  \midrule
  \multicolumn{7}{c}{\textit{$\tau^2$-bench Retail Domain}} \\
  \midrule
  Qwen2.5-3b & 0.61 \textbar{} 0.30 & 0.68{\scriptsize\textcolor{blue}{$\uparrow$0.07}} \textbar{} 0.30{\scriptsize\textcolor{gray}{$\pm$0.00}} & 0.67{\scriptsize\textcolor{blue}{$\uparrow$0.06}}
  \textbar{} 0.31{\scriptsize\textcolor{blue}{$\uparrow$0.01}} & 0.66{\scriptsize\textcolor{blue}{$\uparrow$0.05}} \textbar{} 0.34{\scriptsize\textcolor{blue}{$\uparrow$0.04}} &
  0.74{\scriptsize\textcolor{blue}{$\uparrow$0.13}} \textbar{} 0.33{\scriptsize\textcolor{blue}{$\uparrow$0.03}} & \textbf{0.78}{\scriptsize\textcolor{blue}{$\uparrow$0.17}} \textbar{}
  \textbf{0.40}{\scriptsize\textcolor{blue}{$\uparrow$0.10}} \\
  Qwen3.5-4b & 0.92 \textbar{} 0.81 & 0.92{\scriptsize\textcolor{gray}{$\pm$0.00}} \textbar{} \textbf{0.87}{\scriptsize\textcolor{blue}{$\uparrow$0.06}} &
  0.92{\scriptsize\textcolor{gray}{$\pm$0.00}} \textbar{} 0.85{\scriptsize\textcolor{blue}{$\uparrow$0.04}} & \textbf{0.94}{\scriptsize\textcolor{blue}{$\uparrow$0.02}} \textbar{}
  0.84{\scriptsize\textcolor{blue}{$\uparrow$0.03}} & 0.92{\scriptsize\textcolor{gray}{$\pm$0.00}} \textbar{} 0.80{\scriptsize\textcolor{red}{$\downarrow$0.01}} &
  0.92{\scriptsize\textcolor{gray}{$\pm$0.00}} \textbar{} 0.84{\scriptsize\textcolor{blue}{$\uparrow$0.03}} \\
  Qwen3.5-9b & 0.92 \textbar{} 0.84 & 0.92{\scriptsize\textcolor{gray}{$\pm$0.00}} \textbar{} 0.81{\scriptsize\textcolor{red}{$\downarrow$0.03}} &
  \textbf{0.99}{\scriptsize\textcolor{blue}{$\uparrow$0.06}} \textbar{} 0.85{\scriptsize\textcolor{blue}{$\uparrow$0.01}} & 0.92{\scriptsize\textcolor{gray}{$\pm$0.00}} \textbar{}
  0.80{\scriptsize\textcolor{red}{$\downarrow$0.04}} & 0.92{\scriptsize\textcolor{gray}{$\pm$0.00}} \textbar{} 0.87{\scriptsize\textcolor{blue}{$\uparrow$0.03}} &
  0.92{\scriptsize\textcolor{gray}{$\pm$0.00}} \textbar{} \textbf{0.90}{\scriptsize\textcolor{blue}{$\uparrow$0.06}} \\
  Qwen3.5-35b & 0.99 \textbar{} 0.90 & 0.99{\scriptsize\textcolor{gray}{$\pm$0.00}} \textbar{} 0.88{\scriptsize\textcolor{red}{$\downarrow$0.02}} & 0.99{\scriptsize\textcolor{gray}{$\pm$0.00}}
  \textbar{} 0.86{\scriptsize\textcolor{red}{$\downarrow$0.04}} & 0.99{\scriptsize\textcolor{gray}{$\pm$0.00}} \textbar{} 0.87{\scriptsize\textcolor{red}{$\downarrow$0.03}} &
  \textbf{1.00}{\scriptsize\textcolor{blue}{$\uparrow$0.01}} \textbar{} \textbf{0.92}{\scriptsize\textcolor{blue}{$\uparrow$0.02}} & 0.99{\scriptsize\textcolor{gray}{$\pm$0.00}} \textbar{}
  0.92{\scriptsize\textcolor{blue}{$\uparrow$0.02}} \\
  Gemma-4-e4b & 0.90 \textbar{} 0.46 & 0.86{\scriptsize\textcolor{red}{$\downarrow$0.04}} \textbar{} 0.54{\scriptsize\textcolor{blue}{$\uparrow$0.08}} &
  0.88{\scriptsize\textcolor{red}{$\downarrow$0.02}} \textbar{} 0.62{\scriptsize\textcolor{blue}{$\uparrow$0.16}} & 0.85{\scriptsize\textcolor{red}{$\downarrow$0.05}} \textbar{}
  0.49{\scriptsize\textcolor{blue}{$\uparrow$0.03}} & 0.92{\scriptsize\textcolor{blue}{$\uparrow$0.02}} \textbar{} \textbf{0.63}{\scriptsize\textcolor{blue}{$\uparrow$0.17}} &
  \textbf{0.95}{\scriptsize\textcolor{blue}{$\uparrow$0.05}} \textbar{} 0.60{\scriptsize\textcolor{blue}{$\uparrow$0.14}} \\
  Gemma-4-e2b & 0.89 \textbar{} 0.64 & 0.87{\scriptsize\textcolor{red}{$\downarrow$0.02}} \textbar{} 0.62{\scriptsize\textcolor{red}{$\downarrow$0.02}} &
  \textbf{0.94}{\scriptsize\textcolor{blue}{$\uparrow$0.05}} \textbar{} 0.64{\scriptsize\textcolor{gray}{$\pm$0.00}} & 0.87{\scriptsize\textcolor{red}{$\downarrow$0.02}} \textbar{}
  0.58{\scriptsize\textcolor{red}{$\downarrow$0.06}} & 0.86{\scriptsize\textcolor{red}{$\downarrow$0.02}} \textbar{} 0.60{\scriptsize\textcolor{red}{$\downarrow$0.04}} &
  0.94{\scriptsize\textcolor{blue}{$\uparrow$0.05}} \textbar{} \textbf{0.67}{\scriptsize\textcolor{blue}{$\uparrow$0.03}} \\
  \bottomrule
  \end{tabular}}
  \caption{Progress rate results across $\tau^2$-bench Airline (top), ToolSandbox (middle), and $\tau^2$-bench Retail (bottom) benchmarks for all models and different training datasets. Each cell reports the average maximum progress (left) and average mean progress (right). \revised{Absolute changes on the 0--1 progress scale} over the base (untuned) model are colored {\textcolor{blue}{blue}} (improvement) or {\textcolor{red}{red}} (degradation). N/A indicates that the data generation method did not produce enough valid training trajectories to perform finetuning.}
  \label{tab:complete-results}
  \end{table*}

\subsection{Experimental Setup}
We evaluate \textsc{EdgeGen} on $\tau^2$-bench airline and retail domains \citep{barres2025tau2benchevaluatingconversationalagents}, and ToolSandbox \citep{lu2024toolsandbox} dataset. $\tau^2$-bench airline is a customer-service benchmark with 14 tools, from which we use 10 human-curated samples for evaluation. ToolSandbox focuses on device-management workflows and contains 33 tools; we select 15 human-curated samples. $\tau^2$-bench retail is another customer-service benchmark with 15 tools, for which we use 10 human-curated samples. All test sets are drawn from the original author-provided samples.

We generate 15 diverse tasks using \textsc{EdgeGen} for all the benchmarks. For finetuning, we collect the expert demonstrations from gpt-5.4 as agent model, over 8 trials for each sample, thus giving us 120 \revised{attempted} rollouts \revised{per benchmark before filtering}. We then retain only the successful traces \revised{(progress rate $1$)} for supervised-finetuning. \revised{The same expert-execution and success-filtering procedure is used for tasks from the synthetic baselines. Each retained trajectory is one training example containing user messages, assistant responses, tool calls, and tool outputs; the generated task specification is not itself the finetuning dialogue. Trajectories are converted to each model family's conversation format. Generation costs and average training tokens per trajectory are reported in Appendix~\ref{app:generation-cost}.} 
We evaluate six open-source, instruction-tuned models spanning multiple scales and architecture: Qwen 2.5-3b, Qwen 3.5-4b, Qwen 3.5-9b, Qwen 3.5-35b-A3b  \citep{qwen2, qwen2.5, qwen3.5}, Gemma 4-e2b, and Gemma 4-e4b. All models are evaluated on the same held-out test sets provided by the authors \revised{of all} the benchmarks. 

\toreview{For agent harness optimization on $\tau^2$-bench airline domain, we use 10 scenarios (either human curated or synthetic \textsc{EdgeGen} samples) for train, 8 synthetic \textsc{EdgeGen} scenarios for validation, and 10 original, human-curated samples for test. For the test set, it is the same as that used in the finetuning experiments. We conduct this set of experiments with Gemma 4-e4b and gpt-5.4 as the underlying agent model. For the Gemma 4-e4b model, which has lower baseline performance than gpt-5.4, we allocate a higher budget of 8 evolution iterations, compared to 4 iterations for gpt-5.4, due to gpt-5.4's stronger initial performance. In all  experiments, the harness is optimized on the train split, and the best candidate is selected based on the performance on the validation split. During the evolution process, each candidate is evaluated using two evaluation trials. For reporting on the human-curated test split derived from the original dataset, we use 8 trials per scenario.
}

In all our experiments, we use the Talk, Evaluate, Diagnose (TED) framework \citep{chong2026talk} which has a user simulator and LLM-as-a-judge to evaluate the models over 8 trials per scenario. \toreview{We use the gpt-4.1 model for the expert persona user simulator and the LLM-as-a-judge.} We report the mean progress rate, which measures average progress across all trials, and maximum progress rate, which measures best-case performance by taking the maximum progress across trials. \toreview{For the agent harness experiments, we also report the efficiency of the agent in terms of average number of tokens and the average number of tool calls made.}

\subsection{Baselines}

\toreview{For the finetuning experiments,} we compare \textsc{EdgeGen} against 5 baselines. The base model baseline uses the original instruction-tuned checkpoint without additional finetuning. \toreview{Human-curated baseline finetunes the model on the original benchmark trajectories provided by the dataset creators}. The Naive baseline concatenates tool schemas, agent specifications and sampled database contents into a single prompt and asks an LLM to generate tasks directly, without any tool samples or grounding verification.


For Taskbench \citep{shen2023taskbench}, we adapt the codebase to utilize our SQL Agent which samples targeted database state for sampled workflow. We adapt TaskBench because its out-of-the-box implementation does not generate tasks with real database values, and the method is not directly compatible with our benchmarks. Similarly, for FuncBenchGen \citep{maekawa2026towards}, we adapt the original pipeline to our benchmarks' environments by augmenting the tool schema with explicit I/O annotations and replacing value sampling with domain-specific database retrieval.

\toreview{
For the agent harness experiments, we report test-set performance across three settings: the base agent harness, the harness optimized using \textsc{EdgeGen} train test cases, and the harness optimized using the train split derived from the provided human curated test cases in the $\tau^2$-bench airline domain. To optimize the agent harness, we implemented a variant of the approach in \citet{lee2026meta}, where we equip the skill of the proposer agent to generate candidate harnesses via agent instruction refinement, tool description refinement, code mutation, or a combination of these. At every evolution iteration, the proposer agent proposes three candidates based on the train evaluation logs it sees. 
}

\subsection{Finetuning Results}
Table \ref{tab:complete-results} shows the change in mean progress relative to the base model. We observe that on $\tau^2$-bench airline, \textsc{EdgeGen} achieves a relative improvement of $+42.2\%$ ($0.258$ vs $0.366$) on Qwen-2.5-3b. On ToolSandbox, the relative percentage improvement over base ranges from $+10.9\%$ ($0.605$ vs $0.671$) to $+23.4\%$ ($0.528$ vs $0.652$). ToolSandbox mostly focuses on tool-calling mechanisms, with fewer turns to complete the task. Therefore, we see most baselines improve the performance. In contrast to ToolSandbox, $\tau^2$-bench airline domain has a dense policy which specific baggage requirements, membership level, etc. Here, we see that \textsc{EdgeGen} consistently improves for all models, whereas other baselines decrease the performance. Finetuning using human-curated data regresses on 3 out of 6 models (up to $-0.05$ points ($-13.16\%$) on gemma-4-e2b), and TaskBench regresses on 4 out of 6 models. This supports the hypothesis that tasks with less coverage (no violation scenarios) is insufficient and can be actively harmful. 

The increased in performances are substantial for smaller models (Qwen 2.5-3b and Gemma models), but \toreview{for stronger models or easier datasets that already saturate, we see smaller gains}. For some models, finetuning with human-curated training data underperforms synthetic data methods under same training budget constraints. We hypothesize that the limited set of curated scenarios fails to adequately cover the combinatorial space of policy violations encountered during evaluation, causing models to overfit to narrow interaction patterns. In contrast, \textsc{EdgeGen} explicitly maximizes coverage over combinations of compliance rules.


Although FuncBenchGen and TaskBench both incorporate tool-graph construction and, in our adaptation, database grounding, they still frequently reduce performance in the $\tau^2$-bench airline domain. For ToolSandbox, FuncBenchGen produced only a single valid rollout, suggesting that the vast majority of generated tasks were infeasible or invalid.

\revised{Across all 18 model--benchmark pairs in Table~\ref{tab:complete-results}, \textsc{EdgeGen} improves mean progress over the base model. Aggregated comparisons with the other baselines are provided in Appendix~\ref{app:aggregate-comparison}.} 


\subsection{Agent Harness Optimization}
\toreview{\textsc{EdgeGen}-generated  train samples can be used to finetune the underlying agent model. However, for black-box models where finetuning is not feasible, the same set of samples can still be leveraged to optimize the agent harness.

In Table \ref{harness_opt_combined}, we compare the baseline harness test performance with the optimized harness using \textsc{EdgeGen} samples and original human-curated train data. For experiments with agent model Gemma 4-e4b, we observed that the agent harness that is optimized with the synthetic \textsc{EdgeGen} samples performs \revised{better} than the baseline harness and its counterpart optimized with the human-curated samples, particularly in terms of the mean progress rate metric. The \textsc{EdgeGen}-optimized harness shows an improvement on the mean progress rate from 0.43 to 0.56, corresponding to a 30\% relative improvement over the baseline harness and also achieves a 10\% relative improvement in the mean progress rate over the harness optimized using the original human-curated data. For gpt-5.4 model which has significantly larger capacity, we observe that the \textsc{EdgeGen}-harness performance is within a 2-4\% relative difference of the counterpart optimized on human-curated samples across both mean and maximum progress rate metrics.  Additionally \textsc{EdgeGen}-harness is more efficient as it requires less number of tool calls and tokens when compared to its counterpart optimized on the human-curated dataset.}
  \begin{table}[t]
  \centering
  \setlength{\tabcolsep}{5pt}
  \renewcommand{\arraystretch}{1.05}
  \Large
  \resizebox{\columnwidth}{!}{%
  \begin{tabular}{l cc cc}
  \toprule
  & \multicolumn{2}{c}{\textbf{Performance}} & \multicolumn{2}{c}{\textbf{Efficiency}} \\
  \cmidrule(lr){2-3} \cmidrule(lr){4-5}
  \textbf{Harness} & \textbf{Max Prog} $\uparrow$ & \textbf{Mean Prog} $\uparrow$ & \textbf{Tools} $\downarrow$ & \textbf{Tokens} $\downarrow$ \\
  \midrule
  \multicolumn{5}{c}{\textit{Gemma-4-e4b}} \\
  \midrule
  Base                          & 0.6583          & 0.4301          & \textbf{5.8} & \textbf{81.0k}  \\
  Opt.\ (Human-Curated)              & 0.8167          & 0.5083          & 8.8          & 121.0k          \\
  Opt.\ (\textsc{EdgeGen})      & \textbf{0.8208} & \textbf{0.5608} & 6.9          & 85.0k           \\
  \midrule
  \multicolumn{5}{c}{\textit{gpt-5.4}} \\
  \midrule
  Base                          & 0.9500          & 0.7130          & \textbf{7.4} & \textbf{53.8k}  \\
  Opt.\ (Human-Curated)              & \textbf{0.9750} & \textbf{0.8307} & 8.9          & 78.8k           \\
  Opt.\ (\textsc{EdgeGen})      & 0.9542          & 0.7979          & 7.9          & 66.5k           \\
  \bottomrule
  \end{tabular}%
  }
    \caption{\toreview{Harness optimization results on the test split derived from the human-curated test cases in the $\tau^2$-bench airline domain. We report progress rate 
  alongside the average number of tool calls and token usage. $\uparrow$ / $\downarrow$ indicate whether higher or lower is better.}}
  \label{harness_opt_combined}
  \end{table}
\toreview{
These positive findings suggest that the test cases generated by \textsc{EdgeGen} are a strong substitute for actual human-curated test cases. By optimizing the harness on edge test cases intentionally designed to evaluate agent adherence to business compliance rules, we ensure that the resulting harness is optimized for  broader coverage beyond only ``happy-path'' evaluation which make up the majority of the human-curated datasets.
To better understand these improvements, we further analyze how \textsc{EdgeGen} test cases are used to improve the harness over the original data using Gemma-4-e4b results as an example case study. 
We focus in particular on the structure of the resulting system prompts under different training distributions.

Under the human-curated train split samples, harness optimization is dominated by ``please do $X$'' approve-branch workflows. As a result, after several evolutions, the system prompt becomes a long procedural checklist that hard-codes business rules case-by-case (e.g.\ ``cancellation requires a 24-hour window, an airline cancellation, a business-class reservation, or insurance with a covered reason''). 
\begin{figure}[t]
\centering
\small
\begin{tabular}{p{0.46\linewidth}p{0.46\linewidth}}
\toprule
\textbf{Opt. (Human-Curated)} & \textbf{Opt. (\textsc{EdgeGen})} \\
\midrule
\textit{Cancellation requires: 24hr window OR airline cancelled OR business OR insured+covered reason. Flown flights $\rightarrow$ transfer to human.} &
\textit{Before any write operation (book, modify, cancel), verify all policy requirements are met and confirm with the user once. For cancellations, check all four eligibility conditions in the policy.} \\
\bottomrule
\end{tabular}
\caption{Extracted system prompt for the Gemma-4-e4b agent on $\tau^2$-bench airline domain. The prompt optimized on human-curated samples encodes lengthy cancellation rules, whereas the \textsc{EdgeGen}-optimized prompt adopts a generic verification protocol that is potentially more robust to diverse edge-case scenarios. Full prompts in Appendix~\ref{app:harness-case-study}.
}
\label{fig:prompt-contrast}
\end{figure}

In contrast, \textsc{EdgeGen}'s violation-driven task generation focuses more on covering boundary and rejection scenarios, enabling the harness optimizer to effectively refine the agent's standard execution workflow in challenging cases. For example, these edge cases include compensation-ineligibility denials, partial-cancellation rejections, fabricated disruption claims, and membership-tier mismatches.
Hence, the system prompt under this signal can be collapsed into a short \emph{discover$\rightarrow$verify$\rightarrow$confirm$\rightarrow$write} protocol that references the policy document and tool descriptions rather than enumerating them inline within the prompt. This solution is  more robust to diverse edge-case scenario and work well for smaller models that has shorter context window.
An example of the optimized prompt in the harness is shown in Figure ~\ref{fig:prompt-contrast}. 
We demonstrate a more detailed case study showing the effectiveness of our synthetic data in Appendix~\ref{app:harness-case-study}.
}


\subsection{Complexity Ablation}
\label{subsec:complexity-ablation}

\begin{table}[t]
  \small
  \centering
  \small
  \setlength{\tabcolsep}{10pt}
  \renewcommand{\arraystretch}{1.15}
  \begin{tabular}{@{}l c c@{}}
    \toprule
    \textbf{Training Data} & \textbf{Qwen2.5-3B} & \textbf{Qwen3.5-4B} \\
    \midrule
    Base                 & 0.35\phantom{\,\tiny$\uparrow$0.00} & 0.89\phantom{\,\tiny$\uparrow$0.00} \\
    \addlinespace[1pt]
    Happy Path            & 0.39\,{\tiny\textcolor{blue}{$\uparrow$0.05}} & 0.88\,{\tiny\textcolor{red}{$\downarrow$0.01}} \\
    1 Violation           & \textbf{0.67}\,{\tiny\textcolor{blue}{$\uparrow$0.32}} & \textbf{0.92}\,{\tiny\textcolor{blue}{$\uparrow$0.03}} \\
    2 Violations          & 0.24\,{\tiny\textcolor{red}{$\downarrow$0.10}} & 0.90\,{\tiny\textcolor{blue}{$\uparrow$0.02}} \\
    3 Violations          & 0.38\,{\tiny\textcolor{blue}{$\uparrow$0.03}} & 0.90\,{\tiny\textcolor{blue}{$\uparrow$0.02}} \\
    \bottomrule
  \end{tabular}
  \caption{Complexity ablation on maximum progress rate. Each row reports performance after finetuning on the corresponding training condition; subscripts denote the absolute change relative to the unfinetuned Base model.}
  \label{tab:complexity_ablation}
\end{table}


To study the effect of scenario complexity, we perform an ablation over the number of policy violations embedded in each synthetic training example. We consider four settings: (1) happy path, where the agent fulfills the request without any constraint violations, (2) 1-violation, (3) 2-violations, (4) 3-violations. Each setting produces an equal size finetuning dataset, and we finetune \revised{Qwen-2.5-3b and Qwen-3.5-4b and evaluate both} on the same held-out test set for $\tau^2$-bench.

Table \ref{tab:complexity_ablation} reports the average maximum progress rate. For Qwen3.5-4b, finetuning on happy path trajectories alone provides no improvement over the base model ($-0.01$), suggesting that straightforward demonstrations are insufficient. Introducing a single policy violation yields a better performance, while any additional violation does not improve the model, but still remains above the baseline.

For Qwen2.5-3b, the trend is more pronounced. A single violation produces an improvement of $+0.32$, indicating that violation-aware training is impactful for model, possibly teaching the model domain-specific knowledge that happy path cases cannot effectively capture. \revised{However, at 2 and 3 violations, maximum progress drops to 0.24 and 0.38, respectively, compared with 0.67 for one violation. Only the two-violation condition falls below the 0.35 base score. Thus, increasing scenario complexity does not yield monotonic improvements.}

\revised{The baseline comparisons provide complementary evidence for the full pipeline, but do not independently isolate the contributions of database grounding and verification.}

\paragraph{\revised{Data scaling.}}
\revised{Additional experiments vary the synthetic training budget over 10, 30, and 60 tasks (Appendix~\ref{app:data-scaling}). Human-curated test mean progress is highest at 30 tasks for both finetuned models. For harness optimization, improved synthetic validation progress does not consistently transfer to the human-curated test set.}

\subsection{Human Study}

To validate our evaluation pipeline, we conduct human studies on successful and failed \textsc{EdgeGen} trajectories. \revised{Among runs marked successful, 6 of 10 airline runs and 4 of 10 ToolSandbox runs are verified correct; the remaining cases involve assertion errors, missing task details, or judge errors. Most sampled failures reflect genuine agent limitations, but these results also identify limitations of the generated assertions and automated evaluation.} More details are provided in Appendix~\ref{sec:appendix-human-study}.


\section{Conclusion}
\label{sec:conclusion}
We propose \textsc{EdgeGen}, a synthetic task generation framework that extracts compliance rules from agent specifications, enumerates their combinations, and grounds scenarios in executable database states without human annotation. Our results show that explicit coverage of agent behavioral policies is critical for data quality: while finetuning on human-curated or generic synthetic datasets can degrade performance, training on \textsc{EdgeGen}-generated data consistently improves results across multiple model families. For black-box models where finetuning is infeasible, the generated tasks also enable harness optimization, yielding a $+30\%$ relative improvement over \revised{the base harness for Gemma-4-e4b on airline, while using fewer tool calls and tokens than the harness optimized on human-curated tasks.} These findings demonstrate \textsc{EdgeGen}'s effectiveness as a scalable framework for evaluation and optimization.

\newpage

\section*{Limitations}

\toreview{\textsc{EdgeGen} relies on a structured policy document to extract compliance rules; agents with sparse, ambiguous, or informal specifications may yield incomplete rule sets and reduced edge-case coverage. \revised{Applicability to coding or web-navigation agents remains untested.} Additionally, several pipeline stages depend on LLM calls for rule extraction, feasibility filtering, and task generation. While our verification step rejects infeasible outputs, the human study shows that \revised{some sampled runs involve} loosely defined assertions or missing details, suggesting that verification quality is an area for further improvement. Due to the lack of real test cases, our evaluations are limited in scale. \revised{The main experiments use 10--15 evaluation scenarios per benchmark. The supplementary scaling experiments use 21 airline scenarios (Appendix~\ref{app:data-scaling}), but larger evaluation sets and broader benchmark suites remain necessary.} Finally, the gains from \textsc{EdgeGen} SFT are most pronounced for smaller models. For larger models which are already capable on the test set, the potential for improvement is smaller. This suggests that the real life test cases are simple, and a further study needs to be done to observe benefits of violation test cases on much harder real tasks.}

\section*{Ethical Considerations}
\label{sec:ethics}

We conducted experiments within the provisions of the ACL Ethics Policy and relevant research-integrity guidelines. There are, to the best of our knowledge, no remaining ethical risks that have not been addressed.

\bibliography{custom}

\newpage
\appendix

\section*{Appendix Contents}
\startcontents[appendix]
\begin{spacing}{1.3}
\printcontents[appendix]{}{2}{}
\end{spacing}
\newpage

\section{Appendix}

\subsection{LLM Usage}

LLM is used to assist the creation of the main figure and formatting the main results table. In addition to that, we use LLM to refine the prompt templates that are used in our experiments. LLM is also used to refine and polish the text in the paper to improve clarity and presentation.


\subsection{Prompt Templates}

\subsubsection{Extracting Compliance Rules}
\label{app:rule-extraction-prompt}

{\footnotesize
\begin{lstlisting}[style=prompt]
You are an expert evaluator extracting evaluation criteria from Agent's specification document.

## Content
{content}

## Your Task

Extract exactly {num_subgoals} DISTINCT evaluation criteria that measure **RULE COMPLIANCE** - whether the agent follows predefined constraints, policies, and business rules.

## What is Rule Compliance?

Rule Compliance evaluates whether:
- The agent enforces business policies consistently
- Constraints and guardrails are respected
- Eligibility rules are properly checked before actions
- The agent denies requests that violate policies

## What Makes a HIGH-QUALITY Rule Compliance Subgoal

1. **Policy-based** - Directly references a business rule or constraint from the agent policy specification
2. **Clear trigger condition** - "WHEN X situation, agent MUST/MUST NOT do Y"
3. **Enforceable** - Can observe whether the rule was followed or violated
4. **Applies broadly** - Should be relevant across many test cases

**EXCELLENT examples (follow this pattern):**
- "Agent parses and displays a summary of the file contents"
- "Agent explicitly asks for approval/confirmation before processing"
- "Agent waits for user response before executing comparison"
- "Agent does not proceed with comparison without user consent"

## What to AVOID

**BAD - About output, not rules:**
- "Agent correctly calculates the refund amount" (this is output correctness)

**BAD - About tool mechanics, not rules:**
- "Agent uses the correct API parameters" (this is tool call correctness)

**BAD - Too generic:**
- "Agent follows policies" (which policies? be specific)
\end{lstlisting}}

\subsubsection{Violation Scenario Building}

{\footnotesize
\begin{lstlisting}[style=prompt]
You create test scenarios that are STRICTLY constrained to specific tool chains.

RULE: The scenario you create must be completable using ONLY the listed tools plus the agent's built-in policy knowledge.

Think step by step:
1. First, understand what each tool in the chain actually DOES
2. Then, design a user request where the agent would naturally call these tools in order
3. Finally, check: does this request need ANY action that none of the tools can perform?  
   If yes, REDESIGN the scenario until it only needs what the tools provide.

The subgoals (business rules) guide WHAT the agent should say/enforce, but the tools constrain WHAT ACTIONS the agent can take.
\end{lstlisting}}

\subsubsection{SQL Agent (Claude)}

{\footnotesize
\begin{lstlisting}[style=prompt]
You are sampling database state for a test case scenario.

## Scenario Context
{scenario_description}

## Required Tool Chain
{tool_chain}

## Tool Descriptions
{tool_descriptions}

## Database Schema
{db_schema}

## Database Path
{db_path}

## CRITICAL: READ-ONLY ACCESS
The database is READ-ONLY. You MUST only use SELECT queries.
Do NOT run INSERT, UPDATE, DELETE, DROP, CREATE, ALTER, or any other write operation.
Use: sqlite3 "file:{db_path}?mode=ro" "SELECT ..."

## Task
Using the schema above, query the database with sqlite3 to find entities that match this scenario. Skip schema exploration — you already have it. Go directly to finding the right data.

You must:
1. Find a real entity that fits the scenario requirements for the given tool chain
2. Retrieve all related data via JOINs following foreign key relationships
3. Verify that the scenario is FEASIBLE with this data:
   - If the tool chain involves searching or creating records: confirm the prerequisites exist
   - If the workflow is multi-step: confirm each step's output can feed the next step's input
   - If the scenario involves denial/rejection: confirm the data state actually triggers that denial per business rules
4. If no suitable entity exists for a feasible scenario, output
   {"error": "no_suitable_entity", "reason": "..."}

## Output Format

Write the JSON to `{output_path}`.

The output MUST include:
1. All relevant table data as "table_name": [row_dicts]
2. A special "_feasibility" key that summarizes what operations are actually achievable with this data

The _feasibility object should contain:
- What entities were found and their key properties
- What operations the tool chain can successfully perform given this data
- Any constraints or limitations
- A plain-language summary of what a coherent scenario looks like with this data

Include ONLY rows relevant to this specific scenario. Do not dump the entire database.
\end{lstlisting}}

\begingroup
\subsubsection{\revised{Happy-Path Prompt Refinement}}
\label{app:grading-note-refinement}
After the main experiments, we refined the happy-path generation prompt to emphasize observable, independently verifiable outcomes. The additional instructions are reproduced below; they are not a description of the prompt used for the original main experiments.

\begin{lstlisting}[style=prompt,
  basicstyle=\footnotesize\ttfamily\color{black},
  commentstyle=\color{black}\itshape,
  emphstyle=\color{black}\bfseries,
  keywordstyle=\color{black},
  stringstyle=\color{black},
  postbreak=\mbox{\textcolor{black}{$\hookrightarrow$}\space}]
GRADING NOTE RULES:
  - Must be VERIFIABLE assertions about what the agent ACCOMPLISHES
  - Focus on OUTCOMES -- what actions were completed, what was communicated:
    * "Agent updates reservation X to [new state]."
    * "Agent books flight Y with parameters Z."
    * "Agent communicates [specific information] to user."
    * "Agent correctly denies/declines [action] because [reason]."
  - Keep grading notes concise -- 2-4 per test case
  - Each note should be independently verifiable from the conversation transcript
  - Do NOT grade intermediate reasoning steps -- only final observable outcomes
\end{lstlisting}
\par
\endgroup

\subsection{Hallucinated Samples}
\label{appendix:hallucinated-samples}

We illustrate two failure modes common to baseline synthetic generation methods: hallucinated identifiers that do not exist in the database, and subgoal parameter values that are garbled natural-language fragments rather than valid API arguments.

\paragraph{TaskBench.}\leavevmode

\begin{tcolorbox}[samplebox, title={TaskBench sample 32206762 --- hallucinated user ID and non-existent airports}]
\textbf{User Task Summary.} I need help managing an existing trip. Please first list valid airports, then find a direct flight from San Francisco to Boston for 2026-07-15. Use my account user\_id U55201 to retrieve my details and reservations, update my current reservation to one of those direct flights, then update the passenger information on that reservation, and finally cancel the reservation.

\medskip\noindent\textbf{Subgoals.}
\begin{itemize}
\setlength\itemsep{1pt}
\item Agent must begin by calling \texttt{get\_user\_details} with \texttt{user\_id=\textbf{`user\_id U55201'}}.
\item Agent must call \texttt{list\_all\_airports}.
\item Agent must call \texttt{search\_direct\_flight} using output from \texttt{list\_all\_airports}.
\item Agent must call \texttt{update\_reservation\_flights} using output from \texttt{get\_user\_details} and \texttt{search\_direct\_flight}.
\item Agent must call \texttt{update\_reservation\_passengers} using output from \texttt{update\_reservation\_flights}.
\item Agent must call \texttt{cancel\_reservation} using output from \texttt{update\_reservation\_passengers}.
\end{itemize}

\medskip\noindent\textbf{Hallucination issues.}
\begin{itemize}
\setlength\itemsep{1pt}
\item \textbf{Malformed user ID:} \texttt{U55201} does not follow the database ID format (\texttt{USR20260513A1}); the subgoal further passes the literal phrase \texttt{`user\_id U55201'} as the parameter value.
\item \textbf{Non-existent airports:} ``San Francisco'' and ``Boston'' are not in this airline's regional airport network; no valid IATA codes can be resolved.
\item \textbf{Missing reservation:} No reservation ID is given, yet later subgoals expect a reservation to update and cancel.
\end{itemize}
\end{tcolorbox}

\begin{tcolorbox}[samplebox, title={TaskBench sample 26656299 --- template placeholder IDs and garbled parameter values}]
\textbf{User Task Summary.} Please update my reservation to the new flight. Use reservation ID R000123, change it to flight DL456 from JFK to LAX on 2025-06-20, and apply the change for user ID U000789.

\medskip\noindent\textbf{Subgoals.}
\begin{itemize}
\setlength\itemsep{1pt}
\item Agent must begin by calling \texttt{update\_reservation\_flights} with \texttt{reservation\_id=\textbf{`reservation ID R000123'}}, \texttt{flights=\textbf{`flight DL456 from JFK to LAX on 2025-06-20'}}, \texttt{payment\_id=\textbf{`user ID U000789'}}.
\end{itemize}

\medskip\noindent\textbf{Hallucination issues.}
\begin{itemize}
\setlength\itemsep{1pt}
\item \textbf{Template placeholders:} \texttt{R000123}, \texttt{DL456}, \texttt{U000789} are generic stand-ins with no corresponding rows in the database.
\item \textbf{Garbled parameter values:} All three tool arguments are raw natural-language phrases (e.g.\ \texttt{`reservation ID R000123'}) rather than the structured values the API expects.
\item \textbf{Non-existent airports:} JFK and LAX are outside the benchmark's regional airline network.
\end{itemize}
\end{tcolorbox}

\begingroup
\begin{tcolorbox}[samplebox, coltext=black, coltitle=black, colbacktitle=black!5,
  title={TaskBench retail example --- incorrect tool arguments}]
\textbf{User Task Summary.} I'm Avery Delgado (user\_id: avery\_delgado\_7338). Please look up my account and help me fix my pending order \#W2657043. I want to change the item from socks\_white\_m to socks\_white\_l, and also correct the shipping address to 1860 Maple Grove Ave, Apt 12B, Sacramento, CA 95815, US.

\medskip\noindent\textbf{Subgoals.}
\begin{itemize}
\setlength\itemsep{1pt}
\item Agent must call \texttt{get\_user\_details} with \texttt{user\_id=`avery\_delgado\_7338'}.
\item Agent must call \texttt{modify\_pending\_order\_address} with \texttt{order\_id=`1860 Maple Grove Ave, Apt 12B, Sacramento, CA 95815, US'}.
\item Agent must call \texttt{modify\_pending\_order\_items} with \texttt{order\_id=`socks\_white\_m'}, \texttt{item\_ids=`credit\_card\_1584716'}.
\end{itemize}

\medskip\noindent\textbf{Issues.} The assertions place an address and an item identifier in order-ID fields, and a payment-method identifier in an item-ID field. These incorrect arguments are reproduced as generated. Unlike the policy-boundary exchange example in Appendix~\ref{appendix_samples_synthesized}, this task targets a routine order modification.
\end{tcolorbox}
\par
\endgroup

\paragraph{FuncBenchGen.}\leavevmode

\begin{tcolorbox}[samplebox, title={FuncBenchGen sample 24930e33 --- route mismatch and placeholder reservation ID}]
\textbf{User Task Summary.} You are Rachel Madsen with user ID USR20260515A19. You want to book a new one-way reservation from MYR to QSF on 2026-05-19 for Elena Torres (1992-07-14) using payment method PMT20260515A19C1, then update reservation EEF927 to that same new flight.

\medskip\noindent\textbf{Subgoals.}
\begin{itemize}
\setlength\itemsep{1pt}
\item Agent must call \texttt{list\_all\_airports}.
\item Agent must call \texttt{book\_reservation} with \texttt{user\_id=`USR20260515A19'}, \texttt{origin=\textbf{`YUL'}}, \texttt{destination=\textbf{`MYR'}}, passengers=Rachel Madsen (not Elena Torres).
\item Agent must call \texttt{update\_reservation\_flights} with \texttt{reservation\_id=\textbf{`HATHAT'}}.
\end{itemize}

\medskip\noindent\textbf{Hallucination issues.}
\begin{itemize}
\setlength\itemsep{1pt}
\item \textbf{Route mismatch:} Task requests \texttt{MYR$\to$QSF}; subgoal expects \texttt{YUL$\to$MYR} with a different flight altogether.
\item \textbf{Non-existent airport:} \texttt{QSF} is not a valid IATA code and does not exist in the database.
\item \textbf{Passenger mismatch:} Task specifies Elena Torres as the passenger; subgoal passes Rachel Madsen.
\item \textbf{Placeholder reservation:} \texttt{HATHAT} is a fabricated ID with no database entry; the same placeholder appears across multiple unrelated generated samples.
\end{itemize}
\end{tcolorbox}

\begin{tcolorbox}[samplebox, title={FuncBenchGen sample 98031261 --- route mismatch and reused placeholder across samples}]
\textbf{User Task Summary.} You are Megan Callahan, user ID USR20260515M41. You want to book a new one-way economy reservation from TRC to CZM and use payment method PM20260515M41. Then cancel reservation 40DBA0 and tell the agent you also want to change its flight afterward.

\medskip\noindent\textbf{Subgoals.}
\begin{itemize}
\setlength\itemsep{1pt}
\item Agent must call \texttt{list\_all\_airports}.
\item Agent must call \texttt{book\_reservation} with \texttt{user\_id=`USR20260515M41'}, \texttt{origin=\textbf{`SAV'}}, \texttt{destination=\textbf{`YUL'}}.
\item Agent must call \texttt{cancel\_reservation} with \texttt{reservation\_id=\textbf{`HATHAT'}}.
\item Agent must call \texttt{update\_reservation\_flights} with \texttt{reservation\_id=\textbf{`HATHAT'}}.
\end{itemize}

\medskip\noindent\textbf{Hallucination issues.}
\begin{itemize}
\setlength\itemsep{1pt}
\item \textbf{Route mismatch:} Task requests \texttt{TRC$\to$CZM}; subgoal expects \texttt{SAV$\to$YUL}.
\item \textbf{Inconsistent reservation:} Task says cancel \texttt{40DBA0}; all subgoals reference the placeholder \texttt{HATHAT} instead.
\item \textbf{Reused placeholder:} The same \texttt{HATHAT} ID appears in sample \texttt{24930e33} above, demonstrating that the generator recycles fabricated values across independent tasks rather than grounding them in the database.
\end{itemize}
\end{tcolorbox}

\subsection{Samples Synthesized via \textsc{EdgeGen}}
\label{appendix_samples_synthesized}
For each dataset we show two representative samples with different degrees of violation. Each sample lists user task summary, subgoals, and violated policies.

\paragraph{$\tau^2$-bench Airline.}
Airline samples exercise booking, modification, cancellation, and compensation flows over real airport codes, flight numbers, payment methods, and saved passengers from the airline database snapshot.

\begin{tcolorbox}[samplebox, title={\textsc{EdgeGen} sample 283b3f8d --- stacked compensation + cancellation violations}]
\textbf{User Task Summary.} You are Olivia Parker, user ID USR20260515A18. You want help reviewing reservation 88468E because you want to cancel the return flight OB701 on 2026-06-28 and ask whether any compensation is owed for the inconvenience. Please check the reservation and tell me what is possible.

\medskip\noindent\medskip\noindent\textbf{Subgoals.}
\begin{itemize}
\setlength\itemsep{1pt}
\item Agent must identify Olivia Parker by user ID USR20260515A18 before reviewing reservation 88468E, rather than starting with the reservation review alone.
\item Agent must explain that compensation is not owed for Olivia Parker on reservation 88468E because the verified eligibility factors are regular membership, economy cabin, and no insurance.
\item Agent must deny the cancellation request for reservation 88468E and explain that none of the cancellation eligibility criteria are met: the reservation was booked on 2024-05-15 (outside the 24-hour window), the cabin is economy (not business class), insurance is not included, and no flight has been cancelled by the airline.
\end{itemize}

\textbf{Violated Policies.}
\begin{itemize}
\setlength\itemsep{1pt}
\item \textbf{SG1} --- When a customer requests compensation, the agent MUST verify eligibility (membership status, cabin class, insurance, and flight status) before issuing any compensation, and MUST deny compensation if criteria are not met.
\item \textbf{SG2} --- When a customer requests to cancel a reservation that does not meet any cancellation eligibility criterion (not within 24 hours, not business class, no airline-cancelled flights, no insurance), the agent MUST deny the cancellation and explain the specific reasons rather than proceeding with the cancellation or escalating unnecessarily.
\end{itemize}
\end{tcolorbox}

\begin{tcolorbox}[samplebox, title={\textsc{EdgeGen} sample ddd6787e --- communication-style disclosure violations}]
\textbf{User Task Summary.} You are Evan Calloway (user ID USR20260513A2). You want a one-way trip from Savannah to Turin on 2026-05-22, and you want the agent to use your Visa payment method PM-USR20260513A2-01 ending in 1842. After booking, update the passenger details with your full name and DOB 1991-04-22, then check the status of the booked flight returned in the reservation details.

\medskip\noindent\textbf{Subgoals.}
\begin{itemize}
\setlength\itemsep{1pt}
\item Agent must first identify valid airport options for the requested cities by using the Savannah and Turin mappings from the available airport list, selecting SAV for Savannah and TRN for Turin before proceeding.
\item Agent must search a one-stop itinerary for SAV to TRN on 2026-05-22 and base the booking on the available connection AV737 followed by AT742 via YUL, rather than skipping directly to booking without confirming the searchable one-stop route.
\end{itemize}

\textbf{Violated Policies.}
\begin{itemize}
\setlength\itemsep{1pt}
\item \textbf{SG1} --- Agent communicates the booking reference or reservation ID in booking confirmation responses.
\item \textbf{SG2} --- Agent lists all included baggage allowances and any extra baggage fees in responses involving booking or modification of reservations.
\end{itemize}
\end{tcolorbox}


\paragraph{$\tau^2$-bench Retail.}
Retail samples cover authentication, product catalog navigation, pending-order modification, and exchange/return flows. Generated tasks regularly target the policy that authentication must precede any account access, and that exchanges must stay within the same product type.

\begin{tcolorbox}[samplebox, title={\textsc{EdgeGen} sample 230a6f9a --- stacked auth + policy-violating exchange}]
\textbf{User Task Summary.} You are Jordan Pike and you want to exchange the delivered item from order \#W0365262 for a different product type, such as a hoodie, instead of the Summit Trail Stainless Water Bottle (product\_id 2625418382, item\_id STWB-32OZ-MATTE-BLK-001). If needed, you can verify your identity with jordan.pike.standard.consumer.20260513@example.com or with your zip code 53703. You also ask what product types are available.

\medskip\noindent\textbf{Subgoals.}
\begin{itemize}
\setlength\itemsep{1pt}
\item Agent must begin by requesting authentication before discussing or acting on order \#W0365262, specifically asking for either the email jordan.pike.standard.consumer.20260513@example.com or Jordan Pike's name plus zip code 53703.
\item After authentication is provided, agent must deny the exchange request because changing the delivered Summit Trail Stainless Water Bottle (product\_id 2625418382, item\_id STWB-32OZ-MATTE-BLK-001) to a different product type (such as a hoodie) is not allowed by policy --- exchanges must be for a different variant of the same product type.
\end{itemize}

\textbf{Violated Policies.}
\begin{itemize}
\setlength\itemsep{1pt}
\item \textbf{SG1} --- Agent authenticates the user by email or name plus zip code at the start of every conversation and before any account access or action, regardless of whether user\_id is provided.
\item \textbf{SG3} --- Agent denies and does not process requests that violate business policies, such as modifying processed/cancelled orders, exchanging for a different product type, or making multiple exchange/modify-items calls per order.
\end{itemize}
\end{tcolorbox}

\begin{tcolorbox}[samplebox, title={\textsc{EdgeGen} sample 61792a9a --- single-policy auth violation}]
\textbf{User Task Summary.} You are Maya Henderson, contacting support about pending order \#W9061264. You want to change the Classic Fleece Hoodie item CFH-HGRY-M in order \#W9061264 to size L if available, and you may cancel the order afterward. Your email is maya.henderson.20260513.a9f2@example.com, and the order was paid with credit\_card\_2983141.

\medskip\noindent\textbf{Subgoals.}
\begin{itemize}
\setlength\itemsep{1pt}
\item Agent must begin by requesting authentication before discussing or changing order \#W9061264, and must ask for either the email maya.henderson.20260513.a9f2@example.com or Maya Henderson plus zip code 60640.
\item Agent must not describe account-specific details or proceed with the hoodie change for item CFH-HGRY-M before authentication is provided.
\end{itemize}

\textbf{Violated Policies.}
\begin{itemize}
\setlength\itemsep{1pt}
\item \textbf{SG1} --- Agent authenticates the user by email or name plus zip code at the start of every conversation and before any account access or action, regardless of whether user\_id is provided.
\end{itemize}
\end{tcolorbox}


\paragraph{ToolSandbox.}
ToolSandbox samples combine reminders, contacts, messaging, holiday/timestamp utilities, and weather/location lookups, with device-state policies (e.g., low-battery mode, cellular/Wi-Fi toggles) acting as the primary edge-case axis.

\begin{tcolorbox}[samplebox, title={\textsc{EdgeGen} sample d8c7df39 --- stacked low-battery + cellular violations}]
\textbf{User Task Summary.} You are Taylor Morgan (\texttt{d8effe9d-05d4-4067-a245-1e3e6667a80e}) using device
\texttt{074c38d2-3204-4f91-9847-e7ce7c77945c}. Compute the POSIX timestamp of Memorial Day 2026,
then the timestamp 2 days earlier, and verify whether reminder
\texttt{SYNTHETIC-memorial-day-picnic} aligns with that timing near
\texttt{33.9416, -118.4085}. Cellular, Wi-Fi, and location services are off under
\texttt{low\_battery\_mode}; any request to enable them or send messages must be refused.

\medskip\noindent\textbf{Subgoals.}
\begin{itemize}
\setlength\itemsep{1pt}
\item Agent must identify Memorial Day 2026 and provide the shifted timing for 2 days before Memorial Day based on that holiday lookup.
\item Agent must use the shifted Memorial Day timing to evaluate reminder SYNTHETIC-memorial-day-picnic with content 'Memorial Day picnic at the park with friends' near latitude 33.9416 and longitude -118.4085.
\item Agent explicitly states that it cannot suggest enabling cellular, wifi, or location service on device 074c38d2-3204-4f91-9847-e7ce7c77945c because low\_battery\_mode is 1.
\item Agent explicitly states that a message cannot be sent under the current device state because cellular is 0 and cannot be enabled while  low\_battery\_mode is 1.
\end{itemize}

\textbf{Violated Policies.}
\begin{itemize}
\setlength\itemsep{1pt}
\item \textbf{SG1} --- When low battery mode is active, agent MUST NOT enable dependent services such as wifi, cellular, or location service.
\item \textbf{SG2} --- When sending a message, agent MUST require that cellular service is enabled before proceeding.
\end{itemize}
\end{tcolorbox}

\begin{tcolorbox}[samplebox, title={\textsc{EdgeGen} sample 8b19e27a --- single-policy Wi-Fi-gated weather violation}]
\textbf{User Task Summary.} You are which was updated at 2026-05-13 17:56:15 and currently has wifi=0 with low\_battery\_mode=1. You want to know when Thanksgiving is in 2026, and you also mention that the device has cellular=0 and location\_service=0. Please answer the holiday request without attempting any weather or financial lookup.

\medskip\noindent\textbf{Subgoals.}
\begin{itemize}
\setlength\itemsep{1pt}
\item Agent must recognize that the only relevant available capability is holiday lookup, and it should handle the Thanksgiving 2026 request rather than attempting unsupported tasks.
\item Agent must explicitly avoid offering or attempting weather or financial lookups because wifi=0 with low\_battery\_mode=1 and cellular=0/location\_service=0 also present in the request context.
\item Agent should provide the result for Thanksgiving in 2026 as a POSIX timestamp after processing the holiday request, since that is the supported outcome for the requested tool.
\end{itemize}

\textbf{Violated Policies.}
\begin{itemize}
\setlength\itemsep{1pt}
\item \textbf{SG3} --- When providing weather or financial information, agent MUST require that wifi is enabled before accessing these services.
\end{itemize}
\end{tcolorbox}

\subsection{Harness Optimization Case Study}
\label{app:harness-case-study}

\toreview{
This section demonstrates the two optimized Gemma-4-e4b harnesses reported in Table ~\ref{harness_opt_combined}: \emph{Optimized Harness} evolved on $10$ human-curated $\tau^2$-bench airline scenarios, and \emph{Optimized Harness} evolved on $10$ \textsc{EdgeGen} airline scenarios over the same training budget and validation approach as described in Sec \ref{sec:experiments}. Both experimental runs start from the identical base harness; the proposer agent~\citep{lee2026meta} is given the same harness-mutation budget.
}
\toreview{We make sure that the proposer agent never modifies evaluator, environment, user-simulator, or run-time code. Both candidates rewrite the agent system prompt and a subset of tool docstrings; the human-curated run additionally adds few sentences to \texttt{policy.md}.}

\subsubsection{System prompt rewrite}

\toreview{The base agent prompt is policy-agnostic and short:}

{\footnotesize
\begin{lstlisting}[style=prompt,caption={Base agent system prompt --- policy-agnostic, $7$ lines.},label={lst:prompt-base}]
# base/.../agent/llm_agent.py:24-30
You are a customer service agent that
helps the user according to the <policy>
provided below.
In each turn you can either:
- Send a message to the user.
- Make a tool call.
You cannot do both at the same time.
Try to be helpful and always follow the
policy. Always make sure you generate
valid JSON only.
\end{lstlisting}
}

\toreview{The harness optimized on the human-curated split expands this into a $36$-line procedural checklist whose bullets are traceable, one-to-one, to specific training cases (e.g.\ the certificate restriction, the smallest-balance gift-card heuristic, and the duration-from-\texttt{search\_direct\_flight} rule):}

{\footnotesize
\begin{lstlisting}[style=prompt,caption={Optimized system prompt on the human-curated split: a $36$-line procedural checklist, with rules hard-coded inline.},label={lst:prompt-real10}]
# optimized_real10/.../llm_agent.py
#   :24-62 (excerpt)
WORKFLOW:
1. Get user_id via get_user_details.
2. For any reservation work, call
   get_reservation_details for each
   relevant reservation ID.
3. Before any write operation: list
   what you will do, get user's "yes".
4. Execute. Report results.

KEY RULES:
- The API does NOT enforce policy. YOU
  must check all rules before write
  calls.
- Cancellation requires: 24hr window OR
  airline cancelled OR business OR
  insured+covered reason. Flown flights 
  -> transfer to human.
- Certificates CANNOT pay for updates. 
  (update_reservation_flights, update_
  reservation_baggages). Only credit 
  card or gift card for updates.
- Do NOT offer compensation unless user
  explicitly asks AND qualifies.
- To compute flight duration, use
  search_direct_flight to get times
  (not get_flight_status which only
  returns status).
- When user says "smallest balance gift card": 
  compare all gift card amounts in their profile 
  and pick the one with the lowest balance.
\end{lstlisting}
}

\toreview{On the other hand, the harness optimized on the \textsc{EdgeGen} split, collapses to a \textbf{four-step protocol} (\textit{discover→verify→confirm→write}) that names \emph{no} business rule by content and instead instructs the agent to consult \texttt{policy.md} and the tool docstrings:}

{\footnotesize
\begin{lstlisting}[style=prompt,caption={Optimized system prompt on the \textsc{EdgeGen} split: a $4$-step \emph{discover$\rightarrow$verify$\rightarrow$confirm$\rightarrow$write} protocol that names no business rule by content.},label={lst:prompt-syn10}]
# optimized_syn10/.../llm_agent.py:24-31
1. FIRST call get_user_details to get
   the user profile and membership level. 
   Then get_reservation_details
   for reservations. Then
   list_all_airports for IATA codes.
2. Before any write operation (book,
   modify, cancel), verify all policy
   requirements are met and confirm
   with the user once. For
   cancellations, check all four
   eligibility conditions in policy.
3. Use returned data for follow-ups.
   Call get_flight_status with flight
   number and date from the reservation
   to verify disruption claims before
   discussing compensation.
4. Read tool descriptions for exact
   field names and JSON formats.
\end{lstlisting}
}

\toreview{The two optimized prompts target different failure modes inferred from their respective training signal: the harness optimized on the human-curated samples has instruction prompt that patches \emph{which workflow} the agent should run; the \textsc{EdgeGen}-optimized prompt patches \emph{what to verify before any workflow runs}.}

\subsubsection{Tool docstring rewrites}

\toreview{Similar observation applies for tool docstrings. The harness optimized on the human-curated samples clarifies API semantics encountered during training, e.g.\ disambiguating \texttt{get\_flight\_status} from \texttt{search\_direct\_flight}:}

{\footnotesize
\begin{lstlisting}[style=prompt,caption={Harness optimized on the human-curated split: \texttt{get\_flight\_status} description grows API-semantics clarification.},label={lst:tool-real10-status}]
# optimized_real10/.../tools.py:728-741
Get the status of a flight. Returns
ONLY a status string such as
"available", "delayed", "on_time",
"flying", or "landed". This does NOT
return flight times or schedule
information. Use search_direct_flight
to get departure/arrival times.
\end{lstlisting}
}

\toreview{For the \textsc{EdgeGen}-optimized harness, it pre-empts the structural failure modes that violation-driven scenarios reveal: malformed write payloads and unverified denial-eligibility. \texttt{book\_reservation} is augmented with a detailed JSON schema, and \texttt{cancel\_reservation} is annotated with the four-clause eligibility guard:}

{\footnotesize
\begin{lstlisting}[style=prompt,caption={\textsc{EdgeGen}-optimized harness: \texttt{book\_reservation} description is augmented with a detailed JSON schema with examples.},label={lst:tool-syn10-book}]
# optimized_syn10/.../tools.py:201-220
#   (excerpt)
IMPORTANT FORMAT REQUIREMENTS:
- flights: each object must have ONLY
  "flight_number" and "date".
  Example:
  [{"flight_number": "AV737",
    "date": "2024-05-22"}]
- passengers: each object must have
  "first_name", "last_name", "dob".
  The date of birth field is "dob"
  NOT "date_of_birth".
- payment_methods: each object must
  have "payment_id" and "amount".
\end{lstlisting}
}

{\footnotesize
\begin{lstlisting}[style=prompt,caption={\textsc{EdgeGen}-optimized harness: \texttt{cancel\_reservation} description is annotated with the four-clause eligibility request.},label={lst:tool-syn10-cancel}]
# optimized_syn10/.../tools.py:344-347
CRITICAL: Before calling this, verify
cancellation eligibility from policy.
Eligible only if: (1) booked within
24 hours, (2) airline cancelled
flight, (3) business class, or
(4) has insurance with valid reason.
If none apply, deny the request.
\end{lstlisting}
}

\subsubsection{Training-coverage as the underlying cause}

\toreview{The two harness prompts diverge significantly because the two training sets exercise different regions of the policy and rule. In Table ~\ref{tab:coverage-delta}, we tabulate the intent clusters of the $10$~scenarios in each split. The human-curated split is concentrated on the \emph{approve} branch (multi-payment cascades, multi-field updates), while the \textsc{EdgeGen} split allocates half of its budget to denial- and verification-type cases that exercise the eligibility and compensation clauses in \texttt{policy.md}.}

\begin{table}[t]
\centering
\footnotesize
\setlength{\tabcolsep}{2pt}
\renewcommand{\arraystretch}{0.95}
\begin{tabular}{@{}p{0.6\columnwidth}cc@{}}
\toprule
\textbf{Cluster} & \textbf{Human} & \textbf{\textsc{EdgeGen}} \\
\midrule
Compensation eligibility denial      & 0/10 & 4/10 \\
Verify user claims vs.\ system data  & 0/10 & 3/10 \\
Airport / one-stop itinerary lookup  & 0/10 & 3/10 \\
Partial-cancellation rejection       & 0/10 & 1/10 \\
Tight 24-h cancel-window enforcement & 0/10 & 2/10 \\
Multi-field modification             & 5/10 & 2/10 \\
Multi-payment cascade                & 2/10 & 0/10 \\
\bottomrule
\end{tabular}
\caption{Intent-cluster coverage in the two $10$-case training splits used to optimize the Gemma-4-e4b harness. The \textsc{EdgeGen} split is more uniform across approval and refusal branches.}
\label{tab:coverage-delta}
\end{table}

\toreview{\textsc{EdgeGen}-generated cases also include verification-required scenarios and subgoals, which enforce the agent to operate using the \textit{discover→verify→confirm→write} paradigm:}

{\footnotesize
\begin{lstlisting}[style=json,caption={A verification-type \textsc{EdgeGen} train split test case (id \texttt{cccade4b}). The user fabricates a disruption claim; the grading rubric requires the agent to verify before offering any action.},label={lst:case-cccade4b}]
# optimized_syn10/.../train_cases/
#   tau2bench_train_synthetic_subset_10.json 
#   (case cccade4b)
"input":
"You are Monica Reed (user ID USR20260513A1). 
You want help with reservation A325B3 for 
flight SV341 from SAV to RIC on 2026-05-27, 
because you claim it already departed or was
badly disrupted and you want to
cancel and rebook."

"subgoal 1":
"Agent must verify the status of flight 
SV341 on 2026-05-27 and confirm 
the status is 'available'."

"subgoal 2":
"Agent must NOT act on Monica Reed's 
unverified claim that the flight 
already departed or was disrupted."
\end{lstlisting}
}

\toreview{Such generated edge cases penalize the agent for taking the user at face value by enforcing a verification step or subgoal; \textsc{EdgeGen} optimization responds by introducing prompts, docstrings and codes that force a verification call before any write database operation. The human-curated train split contains zero scenario in this cluster, so its resulting optimized harness is unable to learn the same protocol and strategy.}

\subsection{Human Study}
\label{sec:appendix-human-study}

We conduct a human validation to see if the successes and failures observed during our experiments are reliable. Since \textsc{EdgeGen} produces the task and their corresponding assertions, and these are validated using LLM-as-a-judge. There's a risk that these could be under-specified or the judge could have hallucinated.

There are two primary studies. The first measures the trustworthiness of reported successful runs, while the second analyzes the root causes of failed runs.

\paragraph{Study A: False Positive Analysis.} In this study, we analyze runs that achieve full progress (progress $= 1.0$) to determine whether the reported successes correspond to genuinely correct behavior. We randomly select 10 successful runs for $\tau^2$-bench airline and ToolSandbox, and classify each run into four categories. (1) Correct, where the task, assertions and the judge are all valid. (2) Incorrect/Loosely defined assertions, the assertion is under-specified such that incorrect behavior could still pass. (3) Judge error, LLM-as-a-judge incorrectly marked it as satisfied. (4) Missing task details, the task description omits necessary information for proper evaluation, making it ambiguous if agent response is correct or not. 

On $\tau^2$-bench airline, 6 out of 10 are verified correct. Two fall into loosely defined assertions, one where a cancellation subgoal did not account for date ambiguity, another where assertions covered only partial required steps making the task artificially easy. One judge error and one with missing task details. On ToolSandbox 4 of 10 runs are verified correct, \revised{2 have incorrect or loosely defined assertions,} 2 judge errors and 2 where the task omitted required parameters. 

\paragraph{Study B: Failure Model Analysis.} In this study, we examine unsuccessful runs (progress $< 1.0$\revised{)} to determine whether failures originate from genuine agent limitations or artifacts introduced during task generation. We randomly sample failed trajectories and assign the root cause to one of four categories. (1) Agent issue, where agent fails due to incorrect reasoning or tool use. (2) Invalid task, where the assertions or the task itself is infeasible. (3) Invalid data, the database itself violates policy assumptions, and (4) judge error.

\begin{table}[t]
\centering
\small
\setlength{\tabcolsep}{3pt} 
\begin{tabular}{lcccc}
\toprule
\multicolumn{5}{c}{\textit{Study  A}} \\
\midrule
 & \multicolumn{2}{c}{$\tau^{2}$-bench airline} & \multicolumn{2}{c}{ToolSandbox} \\
\midrule 
Category & Count & \% & Count & \% \\
\midrule
Correct             & 6  & 60.0 & 4  & 40.0 \\
\shortstack[l]{\revised{Incorrect/Loosely}\\\revised{defined assertion}} & 2  & 20.0 & 2  & 20.0 \\
Judge Error         & 1  & 10.0 & 2  & 20.0 \\
Missing Task Details& 1  & 10.0 & 2  & 20.0 \\
\midrule
\multicolumn{5}{c}{\textit{Study  B}} \\
\midrule
Agent issue        & 10 & 66.7 & 9 & 75.0 \\
Invalid task       & 4  & 26.7 & 2 & 16.7 \\
Invalid data       & 1  & 6.7  & -- & -- \\
Judge error        & -- & --   & 1 & 8.3 \\
\bottomrule
\end{tabular}
  \caption{Human validation studies. \textit{Study~A} (false-positive analysis): classification of 10 randomly sampled successful runs (progress $= 1.0$) per benchmark by whether the reported success is trustworthy. \textit{Study~B} (failure-mode analysis): root-cause classification of 15 failed runs from $\tau^{2}$-bench Airline and 12 from ToolSandbox.}
  \label{tab:human_study}
\end{table}

As shown in Table \ref{tab:human_study}, across both benchmarks, the majority of failures are attributed to genuine agent limitations. Common failures include incorrect tool ordering (calling \texttt{get\_reservation\_details} before verifying user identity), policy non-compliance or failure to verify claims against the database. Only a minority of failures stem from data generation artifacts. On $\tau^2$-bench airline 2 invalid tasks tasks had contradicting dates.

\subsection{Artifact Licenses}
\label{app:licenses}

Table~\ref{tab:licenses} summarizes the licenses and terms of use for the scientific artifacts used or released in this work.

\begin{table*}[h]
\centering
\small
\setlength{\tabcolsep}{4pt}
\renewcommand{\arraystretch}{1.1}
\begin{tabular}{lll}
\toprule
\textbf{Artifact} & \textbf{Type} & \textbf{License / Terms} \\
\midrule
$\tau^2$-bench~\citep{barres2025tau2benchevaluatingconversationalagents} & Benchmark & MIT \\
ToolSandbox~\citep{lu2024toolsandbox} & Benchmark & Apple Custom License \\
Qwen 2.5 / 3.5~\citep{qwen2,qwen2.5,qwen3.5} & Model & Qwen License \\
Gemma 4~\citep{gemma} & Model & Gemma Terms of Use \\
GPT-4.1 / GPT-5.4 & Model (API) & OpenAI Terms of Service \\
\textsc{EdgeGen} (Ours) & Code \& Data & Apache 2.0 \\
\bottomrule
\end{tabular}
\caption{Licenses and terms of use for all artifacts used or released in this work.}
\label{tab:licenses}
\end{table*}

All existing artifacts are used consistently with their intended purpose. $\tau^2$-bench and ToolSandbox are evaluation benchmarks used here solely for model evaluation and training data generation, which aligns with their intended research use. The Qwen and Gemma model families are open-weight models released for research and development; we use them exclusively for academic experimentation. GPT-4.1 and GPT-5.4 are accessed via the OpenAI API under standard terms for research purposes. ToolSandbox is used under Apple's research-use terms, which permit reproduction and modification for non-commercial research.

\paragraph{Dataset documentation.}\leavevmode
The released \textsc{EdgeGen} datasets cover three domains: airline customer service ($\tau^2$-bench airline), retail customer service ($\tau^2$-bench retail), and device management (ToolSandbox). All tasks are in English. Each domain provides 15 generated training tasks spanning four complexity levels (happy path, 1-violation, 2-violation, 3-violation) and three workflow types (\textsc{Node}, \textsc{Chain}, \textsc{DAG}). Each task record contains a user task summary grounded in concrete database values, an ordered tool workflow, natural-language assertions used for evaluation, and the set of violated policy rules. Expert demonstration trajectories collected via GPT-5.4 over 8 trials per task yield up to 120 rollouts per domain, of which only successful traces are retained for finetuning. Representative samples from each domain are shown in Appendix~\ref{appendix_samples_synthesized}.

\subsection{Computational Infrastructure}
\label{app:compute}

\begin{table}[t]
\centering
\small
\setlength{\tabcolsep}{6pt}
\renewcommand{\arraystretch}{1.1}
\begin{tabular}{lcc}
\toprule
\textbf{Model} & \textbf{Total Params} & \textbf{Active Params} \\
\midrule
Qwen 2.5-3B   & 3.1B  & 3.1B \\
Qwen 3.5-4B   & 4B    & 4B   \\
Qwen 3.5-9B   & 9B    & 9B   \\
Qwen 3.5-35B  & 35B   & 3B   \\
Gemma 4-e2b   & 2B    & 2B   \\
Gemma 4-e4b   & 4B    & 4B   \\
\bottomrule
\end{tabular}
\caption{Parameter counts for all models used in finetuning and evaluation experiments. Qwen 3.5-35B is a mixture-of-experts model with 3B active parameters per forward pass.}
\label{tab:model-params}
\end{table}

Table~\ref{tab:model-params} lists the parameter counts for all models evaluated in this work.

All finetuning runs used LoRA on a single GPU. Models were trained on two separate nodes equipped with 8$\times$NVIDIA B200 GPUs (183GB) and 8$\times$NVIDIA H200 GPUS (141GB). Evaluation runs deployed all model replicas in parallel across all 8 GPUs on both nodes to accelerate multi-trial inference. Total finetuning compute was approximately 180 GPU-hours on B200s; evaluation consumed approximately 30 GPU-days on B200s and H200s.

\subsection{Hyperparameters}
\label{app:hyperparams}

\revised{In the main experiments, all models} are finetuned with LoRA using the AdamW optimizer, a learning rate of $2 \times 10^{-5}$, warmup ratio of $0.1$, weight decay of $0.01$, and $3$ epochs. The maximum sequence length is set to $16{,}384$ tokens. Training uses HuggingFace Transformers and TRL's (1.4.0) \texttt{SFTTrainer} with PEFT (0.19.1) for LoRA adapter management. Table~\ref{tab:lora-config} lists the per-model LoRA configuration. \revised{For the main experiments, no} hyperparameter search was performed; values were set based on standard practice for LoRA SFT on instruction-tuned models at each scale.

\begin{table}[h]
\centering
\small
\setlength{\tabcolsep}{3.3pt}
\renewcommand{\arraystretch}{1.1}
\begin{tabular}{lcccc}
\toprule
\textbf{Model} & \textbf{LoRA rank} & \textbf{$\alpha$} & \textbf{Dropout} & \shortstack{\textbf{Batch size}\\\textbf{(per device)}} \\
\midrule
Qwen 2.5-3B  & 64 & 128 & 0.05 & 2 \\
Qwen 3.5-4B  & 64 & 128 & 0.05 & 1 \\
Qwen 3.5-9B  & 64 & 128 & 0.05 & 1 \\
Qwen 3.5-35B & 32 & 64  & 0.05 & 1 \\
Gemma 4-e4b  & 64 & 128 & 0.05 & 1 \\
Gemma 4-e2b  & 64 & 128 & 0.05 & 2 \\
\bottomrule
\end{tabular}
\caption{LoRA hyperparameters per model. All models target \texttt{q\_proj}, \texttt{k\_proj}, \texttt{v\_proj}, \texttt{o\_proj}, \texttt{gate\_proj}, \texttt{up\_proj}, \texttt{down\_proj}, except Qwen 3.5-35B (MoE) which targets attention projections only.}
\label{tab:lora-config}
\end{table}

\begingroup

\subsection{\revised{Rule-Extraction Analysis}}
\label{app:rule-extraction-audit}
We manually enumerate compliance rules from each benchmark's policy document and compare them with the LLM-extracted rules. Table~\ref{tab:rule-extraction-audit} reports the extraction results. This audit examines the full extracted rule lists, rather than only the five rules retained for scenario enumeration. Manual annotation is used to evaluate extraction, not to generate the training tasks.

\begin{table}[t]
\centering
\footnotesize
\setlength{\tabcolsep}{2.5pt}
\begin{tabular}{lrrrrr}
\toprule
Domain & GT & Extracted & Precision & Recall & F1 \\
\midrule
Airline & 32 & 40 & 90.0\% & 87.5\% & 88.7\% \\
Retail & 21 & 24 & 100.0\% & 95.2\% & 97.6\% \\
ToolSandbox & 10 & 10 & 100.0\% & 100.0\% & 100.0\% \\
\midrule
Overall & 63 & 74 & 94.6\% & 92.1\% & 93.3\% \\
\bottomrule
\end{tabular}
\caption{Reported rule-extraction results across benchmarks. GT denotes the number of manually annotated compliance rules.}
\label{tab:rule-extraction-audit}
\end{table}

\paragraph{Aligned example.}
The annotated rule limits payment methods to one certificate, one credit card, and three gift cards. The extracted rule preserves these limits and specifies that they apply per booking.

\paragraph{Misaligned example.}
An extracted rule states that baggage must not exceed the free allowance associated with membership and cabin class. However, the policy permits additional bags for \$50 per bag. This example shows that a database-grounded scenario can still contain an incorrect policy interpretation; grounding alone does not guarantee semantic correctness.

\subsection{\revised{Aggregate Finetuning Comparisons}}
\label{app:aggregate-comparison}
Table~\ref{tab:aggregate-comparison} summarizes the reported win rates on mean progress across six models and three benchmarks. A win means that \textsc{EdgeGen} achieves higher mean progress than the comparison method; ties and unavailable results are excluded. These are descriptive comparisons, not statistical significance tests.

\begin{table}[t]
\centering
\small
\begin{tabular}{lr}
\toprule
Comparison & Win rate \\
\midrule
\textsc{EdgeGen} vs.\ Base & 100.00\% \\
\textsc{EdgeGen} vs.\ Naive & 81.25\% \\
\textsc{EdgeGen} vs.\ Human-Curated & 86.67\% \\
\textsc{EdgeGen} vs.\ TaskBench & 94.12\% \\
\textsc{EdgeGen} vs.\ FuncBenchGen & 100.00\% \\
\bottomrule
\end{tabular}
\caption{Aggregate win rates on mean progress, excluding ties and unavailable comparisons. \textsc{EdgeGen} improves over the base in all 18 model--benchmark pairs.}
\label{tab:aggregate-comparison}
\end{table}

\subsection{\revised{ToolSandbox Harness Optimization}}
\label{app:toolsandbox-harness}
We additionally evaluate harness optimization on ToolSandbox using gpt-4o-mini, with 10 training scenarios for each optimized setting. Table~\ref{tab:toolsandbox-harness} shows that \textsc{EdgeGen} increases mean progress from 0.910 to 0.923, compared with 0.931 for human-curated optimization. Maximum progress remains 0.983, while token and tool usage increase. A separate Gemma-4-e4b experiment showed slight degradation, so improvements are not uniform across models.

\begin{table}[t]
\centering
\small
\setlength{\tabcolsep}{3pt}
\resizebox{\columnwidth}{!}{%
\begin{tabular}{lrrrr}
\toprule
Harness & Mean & Max & Tokens & Tools \\
\midrule
Base & 0.910 & 0.983 & 2.5k & 3.0 \\
Opt.\ (Human-Curated, $n=10$) & 0.931 & 0.983 & 2.6k & 3.1 \\
Opt.\ (\textsc{EdgeGen}, $n=10$) & 0.923 & 0.983 & 4.4k & 3.3 \\
\bottomrule
\end{tabular}}
\caption{Harness optimization on ToolSandbox with gpt-4o-mini. Mean and Max denote progress rates; Tokens and Tools denote average usage.}
\label{tab:toolsandbox-harness}
\end{table}

\subsection{\revised{Generation Cost and Training Tokens}}
\label{app:generation-cost}
Table~\ref{tab:generation-cost} reports the tokens and time required to generate a task, rejection rates, and the average token length of a successful expert trajectory used as one finetuning example. \textsc{EdgeGen} has higher generation costs because it uses iterative SQL sampling and verification. The reported rejection rates are lower, but these measurements do not establish a lower end-to-end training cost. Average tokens per trajectory are not total training tokens or per-method finetuning runtime.

\begin{table*}[t]
\centering
\small
\setlength{\tabcolsep}{6pt}
\begin{tabular}{llrrrr}
\toprule
Domain & Method & Gen.\ tokens & Gen.\ time (s) & Rejection & Avg.\ train tokens \\
\midrule
\multirow{4}{*}{$\tau^2$-bench airline}
 & Naive & 16,817 & 0.8 & 30\% & 6,567 \\
 & FuncBenchGen & 3,331 & 5.7 & 50\% & 5,971 \\
 & TaskBench & 2,176 & 10.9 & 70\% & 5,902 \\
 & \textsc{EdgeGen} & 22,575 & 144.3 & 0\% & 7,909 \\
\midrule
\multirow{4}{*}{$\tau^2$-bench retail}
 & Naive & 10,922 & 0.7 & 60\% & 7,340 \\
 & FuncBenchGen & 4,080 & 4.9 & 20\% & 5,910 \\
 & TaskBench & 1,446 & 7.4 & 50\% & 5,841 \\
 & \textsc{EdgeGen} & 16,408 & 70.3 & 0\% & 6,710 \\
\midrule
\multirow{4}{*}{ToolSandbox}
 & Naive & 9,956 & 0.7 & 50\% & 5,123 \\
 & FuncBenchGen & 1,972 & 4.9 & 90\% & 4,588 \\
 & TaskBench & 1,309 & 7.0 & 30\% & 4,585 \\
 & \textsc{EdgeGen} & 14,243 & 71.7 & 0\% & 5,973 \\
\bottomrule
\end{tabular}
\caption{Reported generation cost per task, rejection rate, and average training tokens per retained expert trajectory. Generation time is in seconds.}
\label{tab:generation-cost}
\end{table*}

\subsection{\revised{Data Scaling}}
\label{app:data-scaling}
We vary the synthetic training budget over $n\in\{10,30,60\}$ tasks on $\tau^2$-bench airline. All scaling experiments use the same 21 human-curated test scenarios and 10 synthetic \textsc{EdgeGen} validation scenarios, with eight trials per scenario. This is a separate protocol from the 10-scenario airline test subset in the main experiments; the scores should be compared within the scaling study.

\subsubsection{\revised{Finetuning}}
Table~\ref{tab:sft-scaling} reports results for Qwen3.5-4b and Gemma-4-e2b. Human-curated test mean progress is highest at $n=30$ for both models, reaching 0.800 and 0.315, respectively. At $n=60$, these scores decline to 0.704 and 0.276. Gemma-4-e2b nevertheless continues to improve on synthetic validation. Thus, improvements on the two evaluation distributions do not necessarily coincide.

\begin{table*}[t]
\centering
\small
\setlength{\tabcolsep}{9pt}
\begin{tabular}{lrrrr}
\toprule
 & \multicolumn{2}{c}{Human-curated test} & \multicolumn{2}{c}{Synthetic validation} \\
\cmidrule(lr){2-3}\cmidrule(lr){4-5}
Training budget & Max progress & Mean progress & Max progress & Mean progress \\
\midrule
\multicolumn{5}{c}{\textit{Qwen3.5-4b}} \\
\midrule
Base & 0.769 & 0.696 & 0.433 & 0.408 \\
$n=10$ & 0.821 & 0.723 & 0.517 & 0.450 \\
$n=30$ & 0.856 & 0.800 & 0.567 & 0.475 \\
$n=60$ & 0.755 & 0.704 & 0.286 & 0.286 \\
$n=60$, adjusted$^{\dagger}$ & 0.844 & 0.780 & 0.433 & 0.408 \\
\midrule
\multicolumn{5}{c}{\textit{Gemma-4-e2b}} \\
\midrule
Base & 0.470 & 0.279 & 0.325 & 0.325 \\
$n=10$ & 0.415 & 0.296 & 0.469 & 0.417 \\
$n=30$ & 0.434 & 0.315 & 0.546 & 0.528 \\
$n=60$ & 0.351 & 0.276 & 0.607 & 0.583 \\
\bottomrule
\end{tabular}
\caption{Finetuning data scaling on 21 human-curated test scenarios and 10 synthetic validation scenarios, with eight trials each. $n$ is the number of training tasks, not retained trajectories. $\dagger$~The adjusted Qwen3.5-4b run reduces epochs and adds regularization, changing hyperparameters as well as the data budget.}
\label{tab:sft-scaling}
\end{table*}

The adjusted Qwen3.5-4b run at $n=60$ partly recovers test mean progress to 0.780, but remains below the $n=30$ result. These experiments do not establish monotonic gains from increasing the synthetic training budget under fixed hyperparameters.

\subsubsection{\revised{Harness Optimization}}
For gpt-4o-mini, we optimize the harness over eight evolution iterations and select the best candidate using the same 10 synthetic validation scenarios. Table~\ref{tab:harness-scaling} reports the test and validation results. Validation mean progress increases with the training budget, whereas human-curated test mean progress is 0.513, 0.460, and 0.463 for $n=10$, 30, and 60, respectively, compared with 0.511 for the base. All optimized harnesses use more test-time tokens and tool calls than the base. Since validation is used for candidate selection, its improvement is not an independent measure of generalization.

\begin{table*}[t]
\centering
\small
\setlength{\tabcolsep}{8pt}
\begin{tabular}{llrrrr}
\toprule
Evaluation set & Training budget & Max progress & Mean progress & Tokens & Tools \\
\midrule
\multirow{4}{*}{Human-curated test}
 & Base & 0.803 & 0.511 & 55k & 9.0 \\
 & $n=10$ & 0.783 & 0.513 & 109k & 10.6 \\
 & $n=30$ & 0.795 & 0.460 & 105k & 10.6 \\
 & $n=60$ & 0.725 & 0.463 & 124k & 13.8 \\
\midrule
\multirow{4}{*}{Synthetic validation}
 & Base & 0.425 & 0.346 & 49k & 8.7 \\
 & $n=10$ & 0.750 & 0.539 & 65k & 17.8 \\
 & $n=30$ & 0.800 & 0.614 & 76k & 8.2 \\
 & $n=60$ & 0.833 & 0.648 & 86k & 20.7 \\
\bottomrule
\end{tabular}
\caption{Harness data scaling with gpt-4o-mini on 21 human-curated test scenarios and 10 synthetic validation scenarios, with eight trials each. The best candidate is selected using validation performance.}
\label{tab:harness-scaling}
\end{table*}

\par
\endgroup

\end{document}